\documentclass{article}

\usepackage{arxiv}
\usepackage[utf8]{inputenc} % allow utf-8 input
\usepackage[T1]{fontenc}    % use 8-bit T1 fonts
\usepackage{hyperref}       % hyperlinks
\usepackage{url} 
\usepackage{amsmath} 
\usepackage{booktabs}       % professional-quality tables
\usepackage{amsfonts}       % blackboard math symbols
\usepackage{nicefrac}       % compact symbols for 1/2, etc.
\usepackage{microtype}      % microtypography
\usepackage{lipsum}		% Can be removed after putting your text content
\usepackage{graphicx}
\usepackage{cite}
\usepackage{doi}
\usepackage{authblk}
\usepackage{float}
\usepackage{color}
\usepackage[table,xcdraw]{xcolor}
\usepackage{multirow}
\usepackage{array}
\usepackage{comment}
\usepackage[dvipsnames]{xcolor}
\newcolumntype{C}[1]{>{\centering\arraybackslash}p{#1}}

\hypersetup{
    colorlinks=true,
    linkcolor=blue,
    filecolor=magenta,      
    urlcolor=cyan,
    pdftitle={Overleaf Example},
    pdfpagemode=FullScreen,
    }
\usepackage{wrapfig}

\title{In Context Learning and Reasoning for \\ Symbolic Regression with Large Language Models}

\date{} 					
\author[1]{\normalsize Samiha Sharlin}
\author[1,2]{\normalsize  Tyler R. Josephson}

\affil[1]{\normalsize Department of Chemical, Biochemical, and Environmental Engineering, University of Maryland Baltimore County, \authorcr
1000 Hilltop Circle, Baltimore, MD 21250
}
\affil[2]{\normalsize Department of Computer Science and Electrical Engineering, University of Maryland Baltimore County, \authorcr
1000 Hilltop Circle, Baltimore, MD 21250}

\renewcommand{\shorttitle}{LLMs for SR}

\hypersetup{
pdftitle={A template for the arxiv style},
pdfsubject={q-bio.NC, q-bio.QM},
pdfauthor={David S.~Hippocampus, Elias D.~Striatum},
pdfkeywords={First keyword, Second keyword, More},
}

\begin{document}
\maketitle

\begin{abstract}
Large Language Models (LLMs) are transformer-based machine learning models that have shown remarkable performance in tasks for which they were not explicitly trained. Here, we explore the potential of LLMs to perform symbolic regression -- a machine-learning method for finding simple and accurate equations from datasets. 
We prompt GPT-4 and GPT-4o models to suggest expressions from data, which are then optimized and evaluated using external Python tools.
These results are fed back to the LLMs, which propose improved expressions while optimizing for complexity and loss. 
Using chain-of-thought prompting, we instruct the models to analyze data, prior expressions, and the scientific context (expressed in natural language) for each problem before generating new expressions.
We evaluated the workflow in rediscovery of Langmuir and dual-site Langmuir's model for adsorption, along with Nikuradse's dataset on flow in rough pipes, which does not have a known target model equation.
Both the GPT-4 and GPT-4o models successfully rediscovered equations, with better performance when using a scratchpad and considering scientific context.
GPT-4o model demonstrated improved reasoning with data patterns, particularly evident in the dual-site Langmuir and Nikuradse dataset.
We demonstrate how strategic prompting improves the model's performance and how the natural language interface simplifies integrating theory with data. 
We also applied symbolic mathematical constraints based on the background knowledge of data via prompts and found that LLMs generate meaningful equations more frequently.
Although this approach does not outperform established SR programs where target equations are more complex, LLMs can nonetheless iterate toward improved solutions while following instructions and incorporating scientific context in natural language. 
\end{abstract}

% keywords can be removed

\section{Introduction}
Data analysis is ubiquitous in all disciplines, where identifying correlations between variables is key to finding insights, informing conclusions, supporting hypotheses, or developing a new theory. 
For scientific data, we often aim to find expressions with few adjustable parameters explaining the data while ensuring that they align with theory. 
Symbolic regression is a machine learning technique that approaches equation-based scientific discovery. Given a dataset, it searches through some ``space of possible equations" and identifies those that balance accuracy and simplicity. 
It is different from conventional regression methods, as symbolic regression infers the model structure from data rather than having a predetermined model structure.

Mathematically, symbolic regression is formulated as some form of optimization, not just of the constants in an equation, but as a search through ``equation space'' for optimal expressions. 
In this way, symbolic regression is a form of machine learning – as data is received, an internal model is updated to match the data. When the model fits the data well and can make predictions about unseen data, the algorithm is said to have ``learned'' the underlying patterns in the data. 
In contrast to popular machine learning algorithms like neural networks, symbolic regression not only fits the constants in an equation but also finds functional forms that match the data. 

Symbolic regression methods mainly use genetic algorithms \cite{langleyDatadrivenDiscoveryPhysical1981, kordonApplicationIssuesGenetic2006, schmidtDistillingFreeFormNatural2009, liuImprovementValidationGenetic2016} that generate random expressions from data, optimize their parameters, and evaluate their fitness with respect to the data through an iterative process until a fitness level or a specific number of iterations is reached. Other approaches include using Markov chain Monte Carlo (MCMC) sampling \cite{jin2019bayesian,guimera_bayesian_2020}, mixed integer nonlinear programming \cite{austelGloballyOptimalSymbolic2017, cozadGlobalMINLPApproach2018, cornelio2023combining}, greedy tree searches \cite{de2018greedy}, pre-trained transformer-based models \cite{kamienny2022end, d2022deep}, and sparse matrix algorithms \cite{bruntonDiscoveringGoverningEquations2016, manganInferringBiologicalNetworks2016, ouyangSISSOCompressedsensingMethod2018}. 
These techniques are broadly geared towards accelerating equation search or efficient multi-objective optimization, but they do not integrate reasoning. 
Researchers have long explored ways to make these algorithms more informed by guiding the search space based on the context of the data \cite{goldberg1989cenetic, makarov1998fitting, lu2016using, udrescu2020ai,guimera_bayesian_2020, chakraborty_ai-darwin_2021, kubalik2021multi, engle_deterministic_2022, kronbergerShapeconstrainedSymbolicRegression2022,haiderShapeconstrainedMultiobjectiveGenetic2023, tenachi2023deep, keren2023computational, medina2023active, cornelio2023combining, fox2024}, including using large language models integrated with genetic algorithms \cite{liventsev2023fully,lanzi2023chatgpt,meyerson2023language, bradley2024openelm,guo2024two, hemberg2024evolving, liu2024large, merler2024context,shojaee2024llm, grayeli2024symbolic}. 
As scientific data is strongly tied to theory, encoding it in the program narrows the vast search space and can make the programs more effective.

This work explores an approach to symbolic regression using large language models (LLMs) for equation discovery. 
LLMs are machine learning models adept at understanding and generating natural language. 
At its core, an LLM uses the transformer architecture—a neural network developed by Google that scales very effectively and allows the training of models on massive datasets \cite{vaswani2017attention}. 
The term ``large'' in the language model refers to the size and intricacy of the network, along with the dataset on which it was trained. 
Prior to GPT-3, natural language processing (NLP) tasks were solved by pretraining language models on vast text datasets and fine-tuning them for specific tasks. 
However, GPT-3 demonstrated that language models can excel at tasks using \emph{in-context learning} without necessitating fine-tuning \cite{brown2020language, bommasani2021opportunities}. LLMs are now commonly used for tasks like chat, code generation, summarization, translation, etc. $-$ and quite remarkably, these tasks can be effectively accomplished by using English language as model input without the need for machine learning expertise. 
We have firsthand experience with these models' capabilities by interacting with AI chatbots on platforms like ChatGPT, Claude, or Gemini. 
LLMs have a wide understanding of the world from their training data and can even solve simple math problems expressed in natural language \cite{urrutia2024s, hong2024stuck, satpute2024can}. They are contributing significantly in education and research \cite{kasneci2023chatgpt, kung2023performance, fu2023material,folstad2021future} medicine \cite{coppersmith2014quantifying, mak2023artificial}, physical and social sciences \cite{ferrara2020types, schwaller2022machine, ramos2024review}, as well as in  legal \cite{chalkidis2019neural, guha2024legalbench}, business \cite{bollen2011twitter, goodell2021artificial}, and entertainment \cite{huang2018music, civit2022systematic, singh2021pre, hu2024survey} sectors.

Existing transformer-based symbolic regression programs \cite{valipour2021symbolicgpt,kamienny2022end, d2022deep} use models pretrained on large databases of synthetically-generated dataset/expression pairs, designed specifically for symbolic regression tasks. 
Thus, these approaches learn to pattern-match between datasets and math expressions, but they don't employ iterations like those in genetic programming that optimize expressions for complexity and loss. 
In that respect, LLMs have been used to imitate evolutionary algorithms (EA) \cite{liventsev2023fully,lanzi2023chatgpt,meyerson2023language, bradley2024openelm,guo2024two, hemberg2024evolving, liu2024large}, and have specifically been applied to solve symbolic regression (SR) problems \cite{meyerson2023language,hemberg2024evolving}.
Meyerson and coworkers \cite{meyerson2023language} developed a workflow that performs genetic programming (mutation, crossover, etc.) through prompts in LLMs and tested symbolic regression in two ways: first, by using a language model in all evolutionary operators except the fitness measure, and second, by only using a language model in the initialization, crossover, and mutation operators. 
The second approach more closely resembles our work, but what we do is even simpler: we task LLMs to generate and/or transform expressions freely. 
Furthermore, while these approaches leverage ``in-context'' learning and don't require pretraining \cite{meyerson2023language, hemberg2024evolving}, the context employed is limited to a list of previously-obtained expressions. 
We propose to expand the context to include data, as well as natural language descriptions of the scientific context of the problem. 
To effectively use this context, we anticipate that the LLM will perform better if given time for analysis \cite{austin2021program, chen2021evaluating,wei2022chain}. 
Therefore, we also incorporate zero-shot chain-of-thought prompting with a scratchpad \cite{nye2021show} to frame equation generation for symbolic regression as a reasoning problem \cite{llm_prompting_blog} in the context of the data and free-form scientific information. 

Symbolic regression requires equation generation \emph{and} precise fitting of numerical constants. Yang and coworkers \cite{yang2024largelanguagemodelsoptimizers} show that LLMs can perform linear regression, optimizing constants in math expressions via feedback loops, without revealing the LLM the analytical form. 
They find that `LLM can often
calculate the solution directly from the analytic form,' however, we find LLMs to be unreliable and inefficient for such tasks \cite{frieder2023mathematical}. Consequently, we interleave LLM-based optimization with gradient-based optimization, following a similar approach as \cite{guo2024two}, iteratively refining prompts for more accurate output.
In our work, an LLM guides optimization of the symbolic structure of the math expressions, while SciPy performs numerical optimization of the constants. 
This approach aligns closely with two recent works \cite{merler2024context,shojaee2024llm} in the literature where LLMs were used to generate expressions and optimizations were performed via iterations. 
Although there are some variations in the workflow and significant differences in the explored datasets, the unique contribution of this work lies in using data, context, and theoretical constraints in natural language as model input and implementing a scratchpad to record raw model output, as it solves the problem. 
This promotes informal reasoning about the data and context, and helps reveal ``test set leakage.''  

\section{Methods}
\subsection{Preliminary tests}

In our initial tests, we asked LLMs to generate mathematical expressions from scientific data. 
We used a simple prompt to assess its capability and tested GPT-3.5-turbo and GPT-4 at varying temperatures. 
In all cases, the models produced expressions while ``hallucinating'' arbitrary coefficients. 
We revised the prompts to ask the LLMs to ``show all steps'' to gain insight into how the model selects these values \cite{wei2022chain}. 
In response, the output either provided Python code for optimization or at times, mathematical steps for optimization, but with the wrong output.

\begin{figure}[H]
    \centering
    \includegraphics[width=0.7\textwidth]{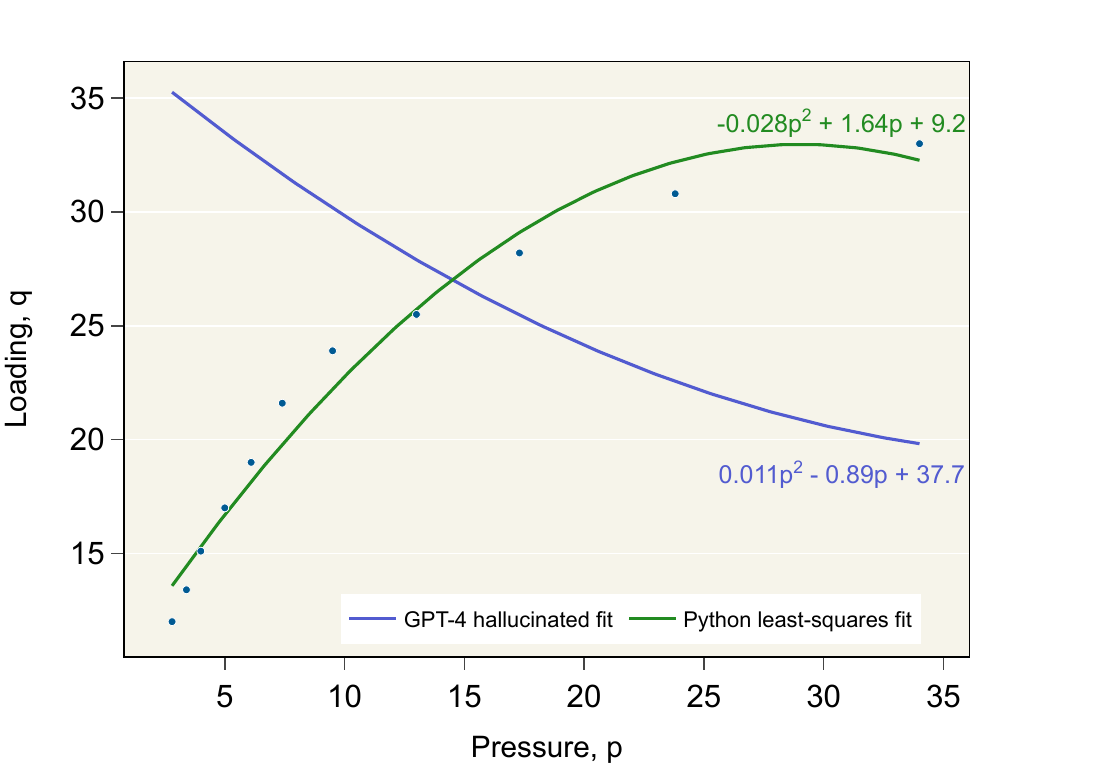}
    \caption[GPT-4 hallucinating curve fitting]
    {\textbf{GPT-4's hallucinated curve fitting on adsorption data.} The plot shows how GPT-4 predicts expressions with coefficients when passed a dataset for nitrogen adsorption on mica \cite{langmuir_adsorption_1918}. The Python code that it generates is correct; however, the actual optimized parameter values produced by running the code differ from what GPT-4 outputs. Figure \ref{gpt3.5Test1} and \ref{gpt4Test1} in SI show the raw output from GPT-3.5  turbo and GPT-4.} 
    \label{hallucination}
\end{figure}

As illustrated in Figure \ref{hallucination}, although the LLM-generated code runs and accurately performs curve-fitting, the model still hallucinates incorrect coefficients as its final output.
Nonetheless, the generated remarks about the data patterns suggested that they may have the potential to generate accurate functional forms that can be optimized outside the LLMs.

\subsection{System design}

\begin{figure}[H]
    \centering
    \includegraphics[width=0.85\textwidth]{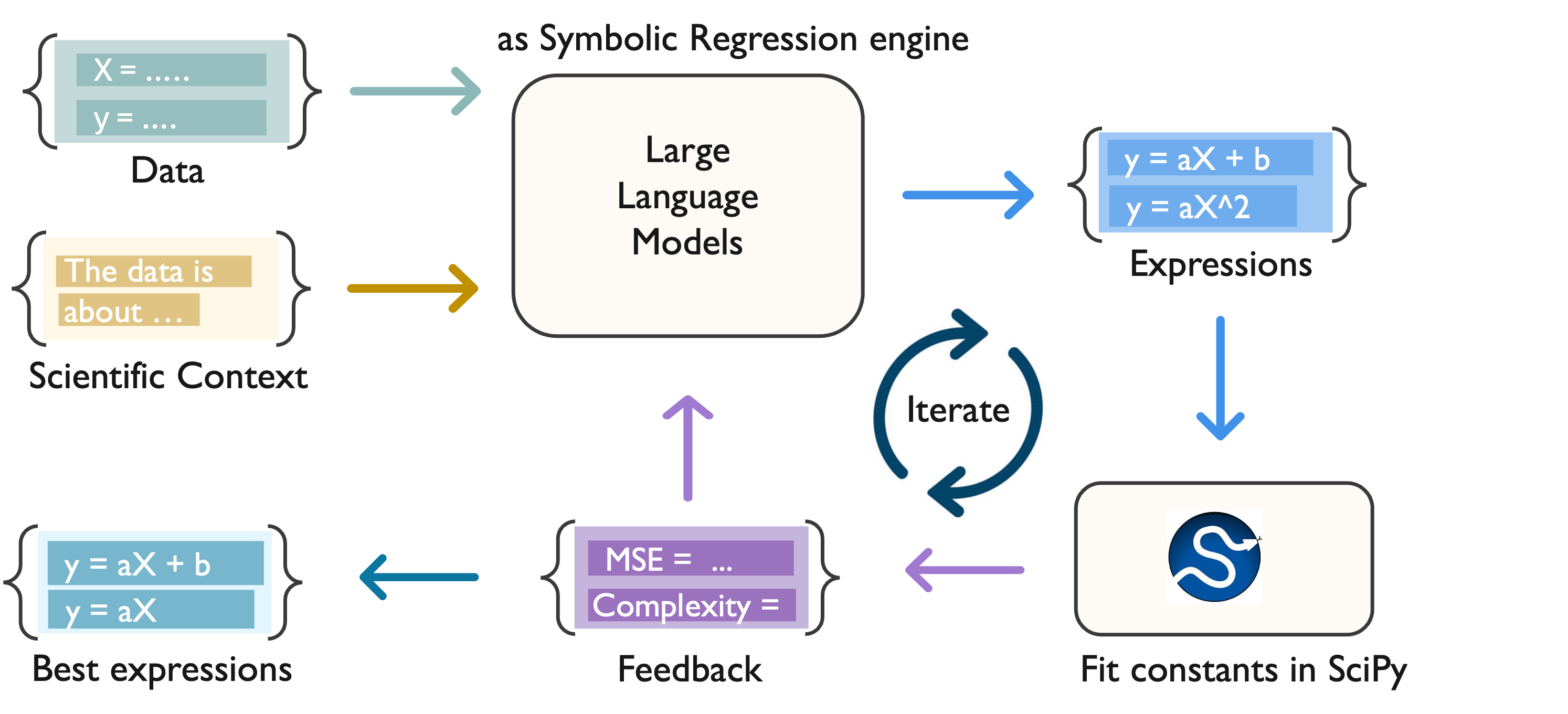}
    \caption[LLM-SR workflow]
    {\textbf{Workflow for using LLMs for SR.} First, the dataset is sent to LLMs (with or without context), which are instructed to suggest expressions without optimizing parameters. The generated expressions are then parsed by Python and optimized using SciPy (Nelder-Mead \cite{nelder1965simplex} method with basin-hopping as the numerical optimizer \cite{wales1997global}). Results for each expression are stored in a Python dictionary and added to a list of results from prior expressions. The top-performing expressions are sent as feedback to LLMs, which are asked to suggest better expressions that optimize for both complexity and loss. The feedback loop is run for a set number of iterations.}
    \label{gptWF}
\end{figure}

Therefore, we designed a workflow (Figure \ref{gptWF}) where we task models GPT-4 and GPT-4o with suggesting expressions without fitting constants, and subsequently, we optimize the coefficients of the expressions using SciPy outside the LLM. 
A Python class takes in expressions, optimizes them, and then calculates their complexity and mean squared error (MSE). 
The results are then stored as a dictionary, the text of which is passed back to LLMs in a subsequent prompt asking to suggest better expressions. 
We initially evaluated GPT-3.5-turbo, but found it less reliable in following instructions than GPT-4 and GPT-4o, more frequently generating expressions that didn't parse. 
Although LLMs could have been used to generate the curve-fitting code, we ensure reliability and prevent reward hacking by writing and validating the code ourselves.

\begin{wrapfigure}{r}{0.4\textwidth}
    \centering
    \includegraphics[width = 0.75\linewidth]{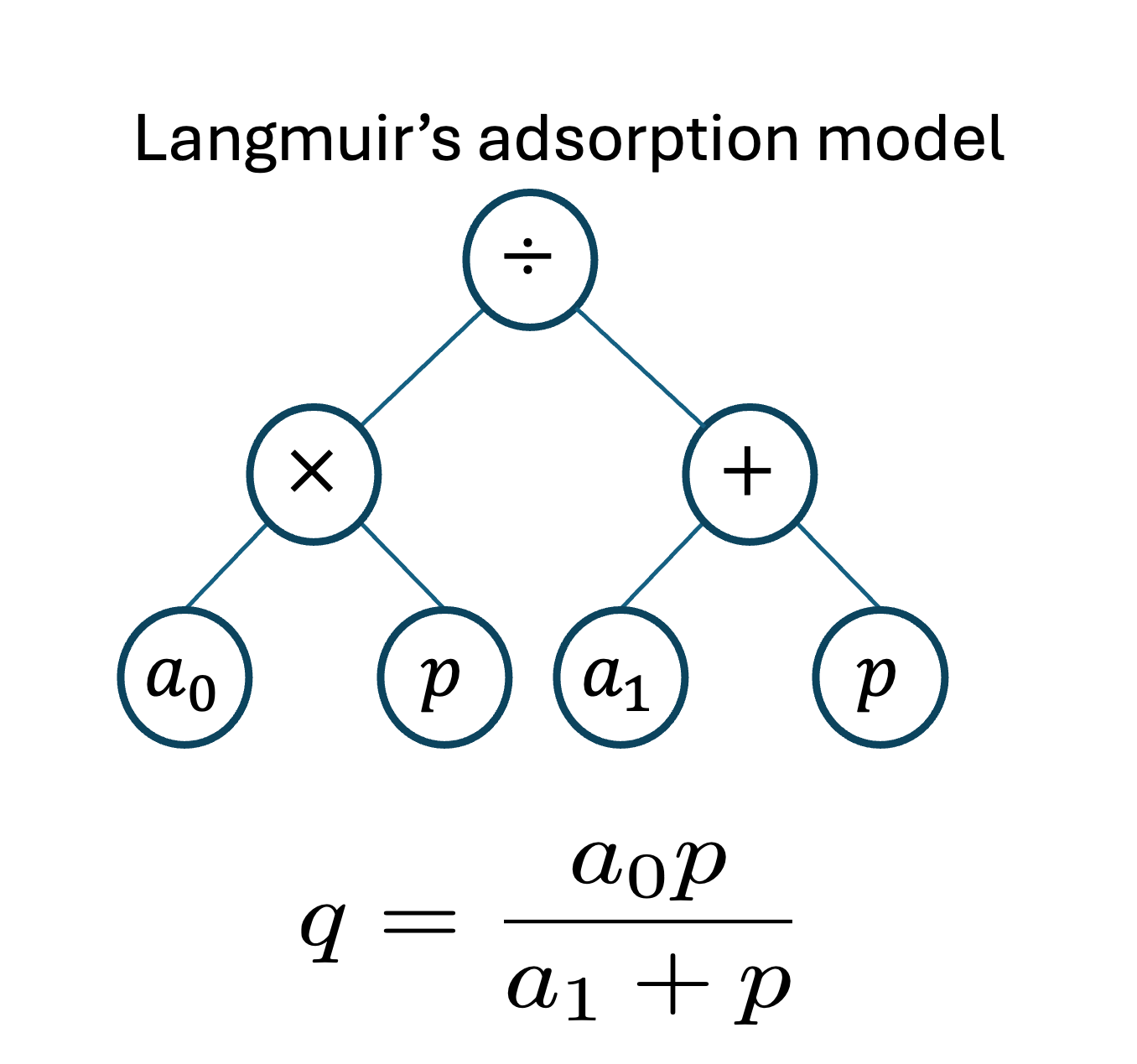}
    \caption[Expression tree for equation complexity]
    {\textbf{Langmuir's adsorption model illustrated in expression tree.} The equation `complexity' is computed from the equation string: each binary operator (+, -, *, /, **) is counted and weighted by 2, each occurrence of a unary function name (e.g. log, exp, sqrt, sin, etc.) is counted with weight 1, and a base cost of 1 is added so that even a constant expression scores at least 1.} 
    \label{langmuirModelTree}
\end{wrapfigure}

We define complexity as the number of operators and variables/constants in a given expression, or in other words, the total nodes in an expression tree (see Figure \ref{langmuirModelTree}). 
A mathematical expression can be represented as a rooted tree structure where internal nodes correspond to operators, which can be binary (e.g., $+$, $*$, $/$, or \char`^) or unary (like sin or log). 
The leaves, or terminal nodes, represent either variables (e.g., $x_1$, $x_2$) or constant values (e.g., $c_1$, $c_2$). 

We use two prompts: 1) Initial Prompt - where we input data and ask GPT-4 and GPT-4o to suggest expressions, and 2) Iteration Prompt - where, along with data, we provide feedback in JSON format. We sort the expressions in descending order of mean squared error, then share this list as feedback, asking the LLMs to suggest new equations optimizing for complexity and loss. In addition to this, we also include a system message to guide the behavior of the language model, setting the tone of the conversation. We do not use chat history or any advanced forms of memory \cite{bae2022keep,xu2021beyond, xu2022long,zhong2022less} in this workflow. The history of generated expressions is maintained externally and provided as feedback in the iteration prompt.
A Python function maintains this feedback loop by sorting and filtering a list of dictionaries based on MSE and complexity. At the start of the search, up to 6 expressions are always returned; later, the least accurate expressions that are not on the Pareto front are pruned in order to manage the length of the context window. By making each call to LLMs independent, we aim to minimize hallucinations that have been observed when the chat history becomes too large in conversational models \cite{liu2024lost, shi2023large}, as well as manage cost by pruning the large quantity of generated scratchpad text.

We can get the most out of an LLM by providing strategic text in the prompt. Prompt engineering is an art that involves structuring prompts that guide the model toward generating desired outputs. However, there is no one-size-fits-all method for crafting optimal prompts, as outcomes depend on the specific task and model.
Even minor changes in wording or structure can influence the model's output.
Various prompting guidelines have been explored in the literature \cite{liu2023pre, reynolds2021prompt, qin2021learning, lester2021power}, including providing clear instructions, emphasizing relevant context using text delimiters, breaking tasks into multiple steps, and incorporating problem-solving conditions. We implemented strategies like incorporating examples and a scratchpad and subsequently refined the prompts.

\subsection{Prompt engineering}
We prototyped our system by testing its ability to rediscover the Langmuir adsorption isotherm ($q = c_1 * p / (c_2 + p)$) from experimental data \cite{langmuir_adsorption_1918}.
This enabled us to quickly identify major structural improvements to the workflow; we further tailored the prompts while testing on more difficult problems.

\begin{figure}[H]
    \centering
    \includegraphics[width=0.7\textwidth]{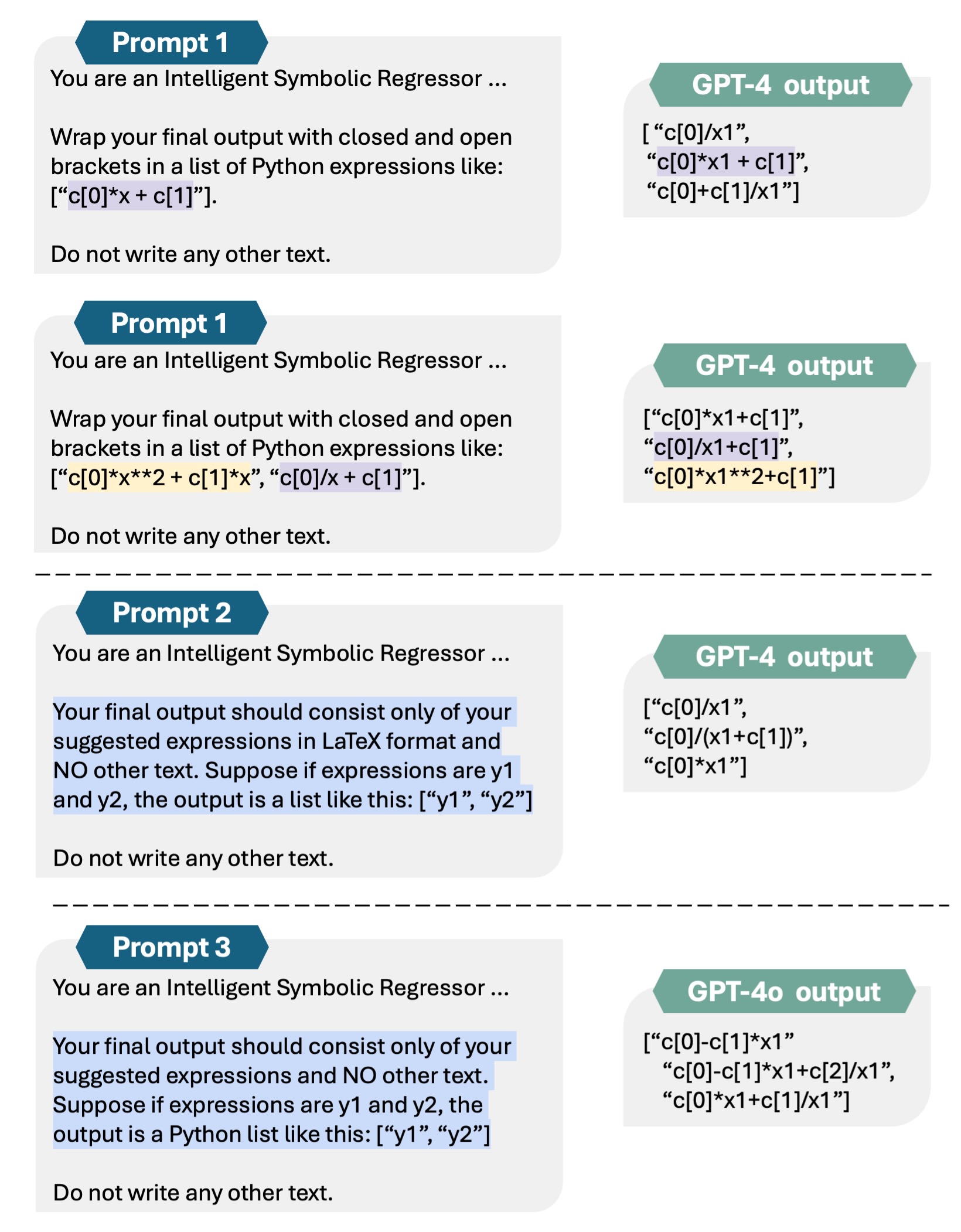}
    \caption[Bias from prompt examples]
    {\textbf{Bias from prompt examples} GPT-4 generated exact expressions from the prompt examples that were given to illustrate the output syntax. The revised prompts (Prompt 2 and 3) eliminate this bias from the equation prediction process.}
    \label{exampleBias}
\end{figure}

\indent \textbf{Removing bias:} We aimed to make the workflow run smoothly without any human intervention, and therefore, it was important to obtain machine-readable and precise output from LLMs to ensure the SciPy function runs without any errors. A simple way to illustrate the expected outcome was by providing examples in a few-shot context \cite{brown2020language, srivastava2022beyond}. While this led to expressions matching the required syntax, we noticed the generated expressions resembled the examples we provided (Figure \ref{exampleBias}). While this taught the LLM correct syntax, it introduced bias that severely compromised the search.

This motivated our two-prompt setup, with an initial prompt tasking LLMs to generate unbiased expressions in LaTeX and an iteration prompt receiving previously-generated examples as Python script (now without bias).
To bridge these, we converted the LaTeX text into SciPy-compatible text using a Python function for string formatting, which we developed after identifying the types of expressions GPT-4 was prone to generating. 
Although this approach did not completely resolve syntax errors, it effectively managed them for the GPT-4 model. While running the GPT-4o model, we realized that most syntax errors stemmed from incorrect LaTeX formatting. To address this, we made a minor adjustment by removing the term "LaTeX" and allowing the model to generate expressions as Python strings instead. This small change was sufficient to remove bias and was also less prone to syntax errors. We nervetheless processed the generated Python expressions through our formatting function, and it worked seamlessly. 

\begin{figure}[H]
    \centering
    \includegraphics[width=1\textwidth]{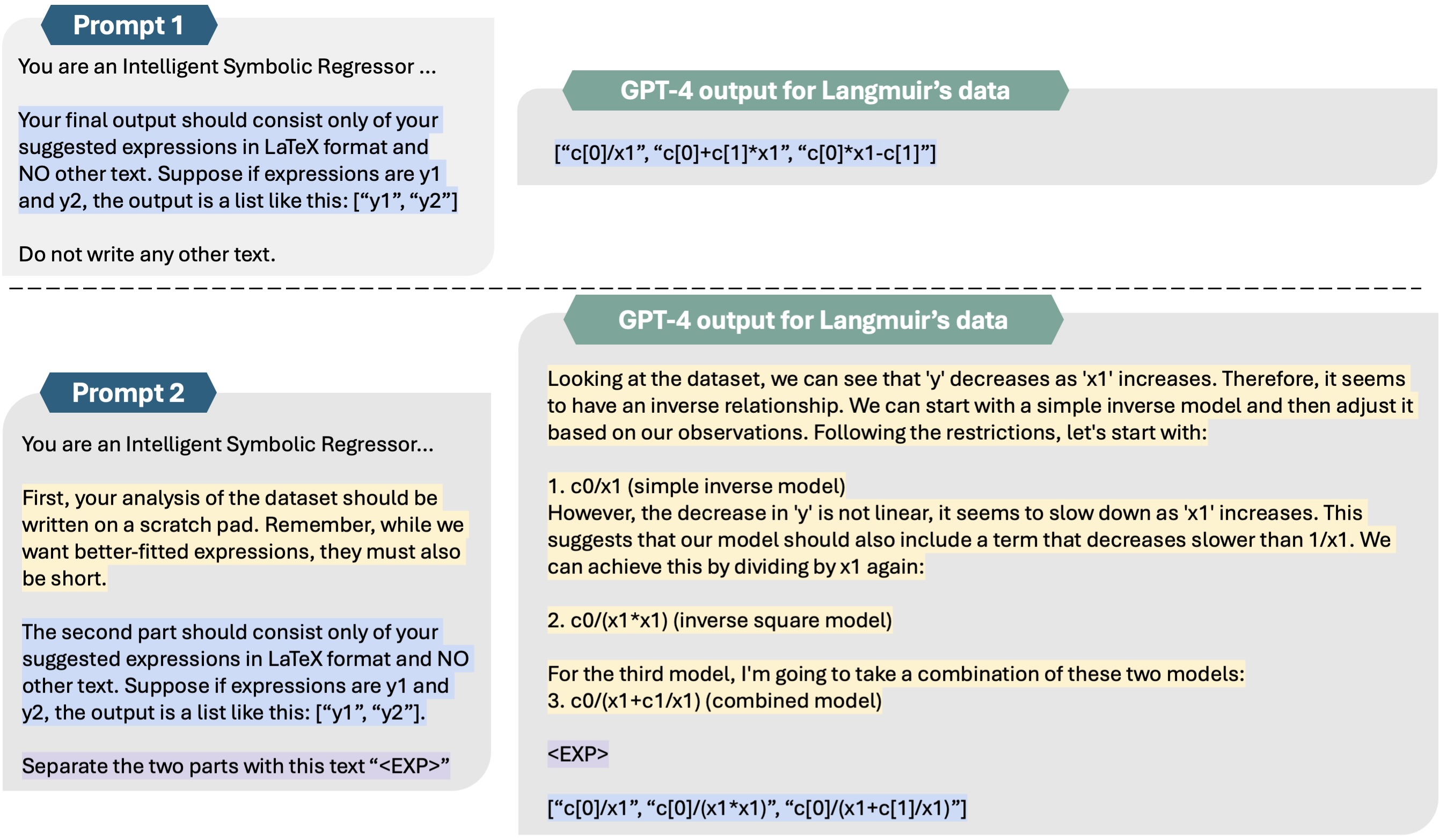} 
    \caption[Effect of scratchpad]
    {\textbf{Effect of scratchpad on Langmuir Dataset.} We observe substantial and qualitative improvements in the predicted expressions after implementing the scratchpad technique (Prompt 2). The suggested expressions for Langmuir's dataset include operators ($/$) present in the target model ($y = \frac{c_1*x}{c_2+x})$.}
    \label{scratchPad}
\end{figure}

\indent \textbf{Recording analysis in a scratchpad:} Studies have shown that LLM performance can be improved by slowing down the model or breaking down its tasks into smaller steps \cite{austin2021program, chen2021evaluating}. One popular strategy is the ``scratchpad" technique \cite{nye2021show}, which mimics how we solve problems by jotting down notes before presenting a final answer in exams.
We implemented this in our workflow, instructing LLMs to generate responses in two parts: data analysis and observations in a scratchpad, followed by its conclusions. After implementing this technique, the model immediately generated higher-quality expressions (Figure \ref{scratchPad}).

\begin{figure}[H]
    \centering
    \includegraphics[width=1\textwidth]{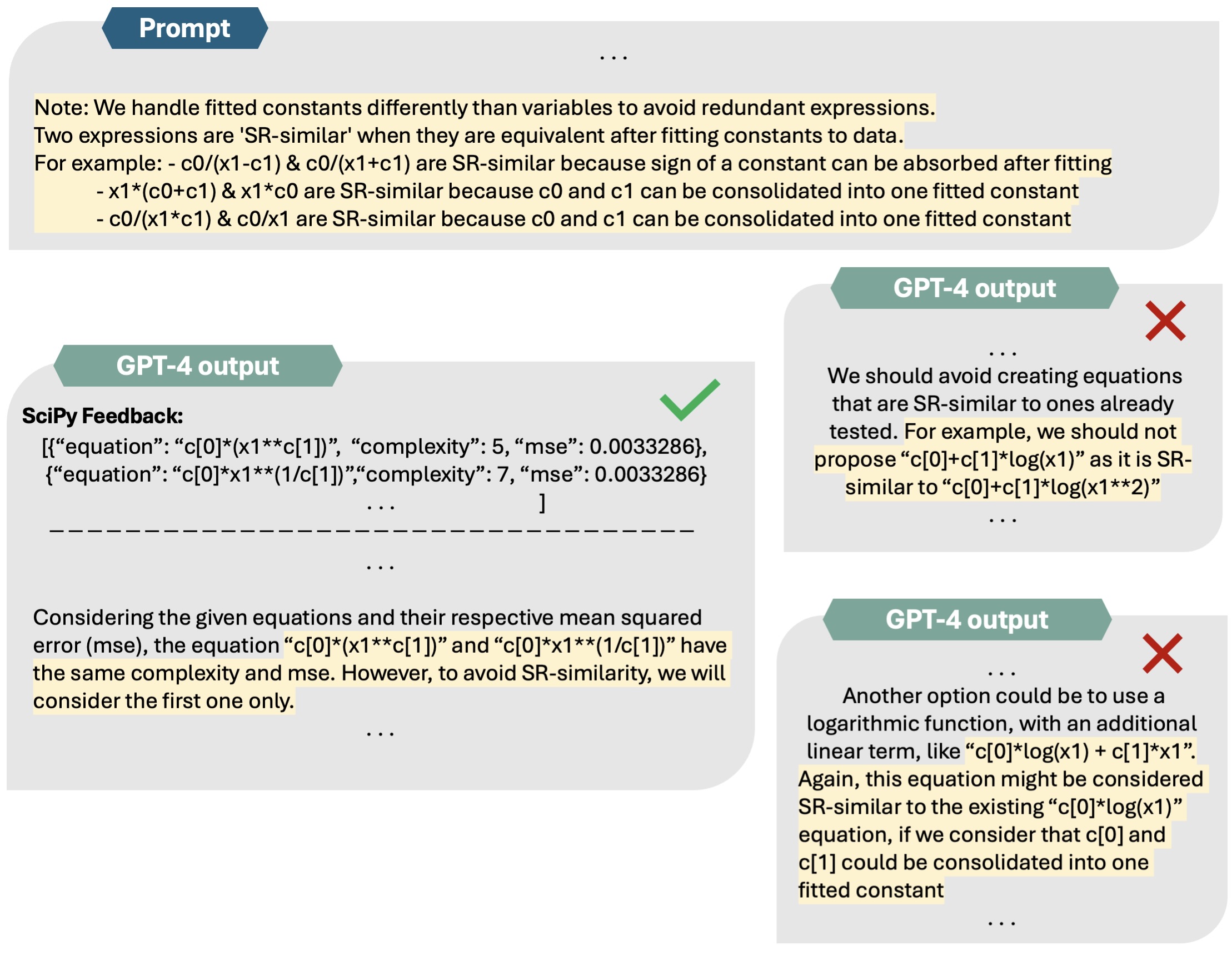}
    \caption[GPT-4 outputs for SR-similar expressions]
    {\textbf{Illustration of GPT-4 outputs for SR-similar expressions.}}
    \label{SR_similar}
\end{figure}

\indent \textbf{Preventing redundant expressions:} The LLMs often generated expressions like $x+c_1$ and $x-c_1$, implying they are different. However, since the constants are yet to be fitted, these expressions are the same from a symbolic regression perspective.
While a computer algebra system like SymPy \cite{sympy} could in principle catch some redundant expressions by simplification to a canonical form \cite{guimera_bayesian_2020, fox2024}, this wouldn't distinguish ``SR-similar" expressions that become equivalent after fitting constants.
Instead, we used prompt engineering to guide generation toward unique expressions: we added a note in the iteration prompt with examples showing how expressions in symbolic regression are similar before parameters are optimized.
While this didn't completely resolve the issue, we did observe a reduction in occurrences, and at times, the scratchpad revealed the models would correctly address this by taking the examples into account (see Figure \ref{SR_similar}).

\begin{figure}[H]
    \centering
    \vspace{-1em}
    \includegraphics[width=1\textwidth]{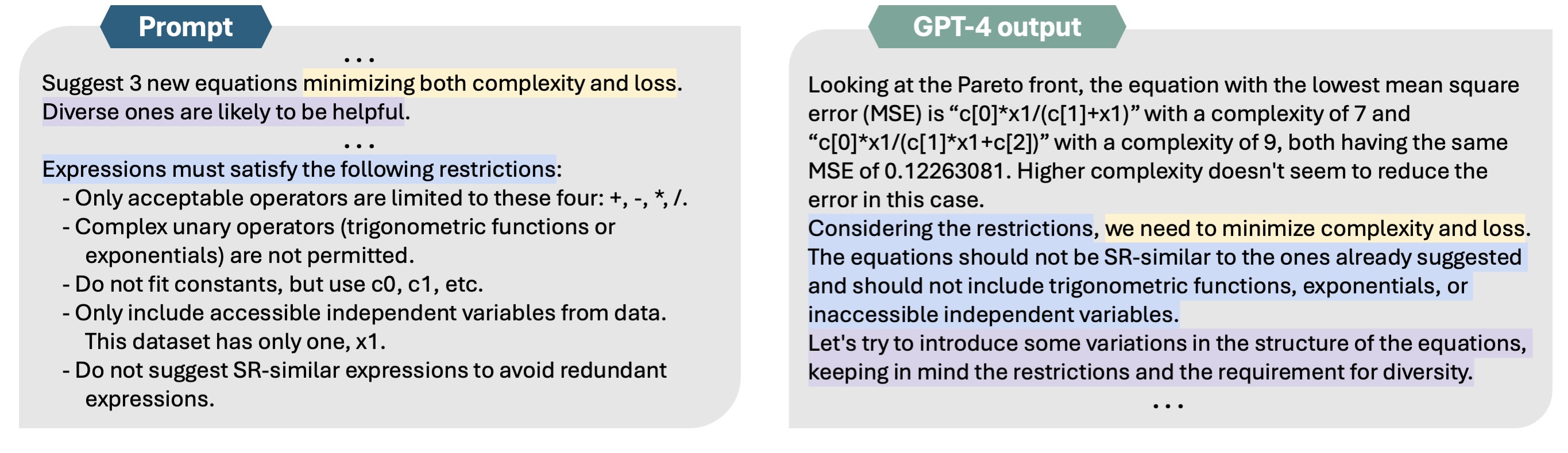}
    \caption[GPT-4 following prompt restrictions]
    {\textbf{Illustration of GPT-4 following restrictions from prompt.}}
    \vspace{-1em}
    \label{promptInstruct}
\end{figure}

\indent \textbf{Avoiding uninteresting expressions:} During the iterative runs, LLMs attempted to improve its accuracy by repeatedly adding linear terms to suggested expressions from previous iterations.
To address this issue, we encouraged the model to explore diverse expressions in the prompt. 
Additionally, in cases involving datasets with multiple independent variables, LLMs would sometimes recommend excluding variables that exhibited weak correlation with the overall dataset pattern.
While this may be useful in some contexts, we wanted expressions that made use of all of the available data, 
so we explicitly instructed the use of all variables.
Additional constraints we implemented included limiting the types of math operators to include and preventing generation of implicit functions, as shown in Figure \ref{promptInstruct}. 

\indent \textbf{Consider scientific context:} 
Our primary motivation for building this system was to test whether providing scientific context could shape the expressions generated by the LLM. SR programs often successfully generate expressions that fit the data well and are simple, yet they may not adhere to scientific principles or be otherwise ``meaningful.''
Yet scientists often have valuable insights into their domain that extend beyond these constraints, and they may not always know which specific expressions will best capture the nuances of their dataset or if entirely new expressions might be more effective. 
By incorporating scientific context, we aim to align and constrain the equation search to be consistent with scientific theories.

\begin{figure}[H]
    \centering
    \includegraphics[width=1\textwidth]{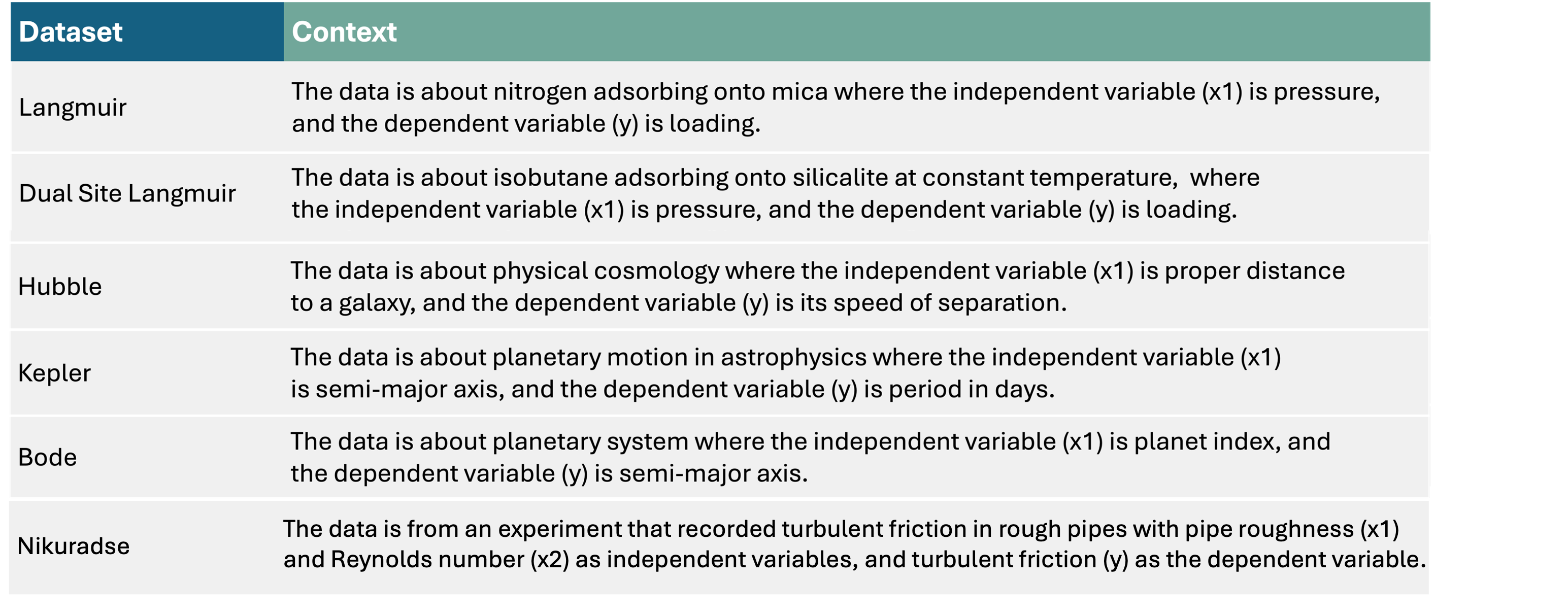}
    \caption{\textbf{Context provided to LLMs for all the datasets}}
    \label{contextTab}
\end{figure}

Classical SR approaches always incorporate some amount of guidance from the practitioners (e.g. by limiting available math operators and variables and incorporating some bias toward parsimonious expressions); more recent work enables SR programs to explicitly account for limiting behavior (shape-constrained SR) and dimensional constraints \cite{schmidt2009incorporating, arnaldo2015building, de2012deap, haiderShapeconstrainedMultiobjectiveGenetic2023, lu2016using, tenachi2023deep, brence2023dimensionally}.
Incorporating these into SR algorithms typically requires bespoke modification of research software; our strategy is to use LLM prompting in natural language to instill this context, which might additionally include more fluid constraints like ``generate diverse equations'' and ``consider scientific context.''
%For instance, the context we provide for Kepler's Law is a single line text stating, ``The data is about planetary motion in astrophysics where the independent variable (x1) is semi-major axis, and the dependent variable (y) is period in days." 
Figure \ref{contextTab} in SI lists the context provided for each of the datasets. 

\section{Results}
We focused on applying LLM-based SR to experimental datasets, exploring its potential to uncover interpretable scientific relationships rather than benchmarking on synthetic datasets.
Specifically, we investigated two well-known chemistry datasets: Langmuir and dual-site Langmuir adsorption isotherm models \cite{langmuir_adsorption_1918}.
These two datasets were also examined in our previous work \cite{cornelio2023combining, fox2024}, which enables us to directly compare this LLM-based SR approach with traditional SR methods. 
We further tested our workflow on a challenging dataset of friction losses in pipe flow \cite{nikuradse1950laws}, which does not have an established target model and has been the subject of prior studies by other SR programs \cite{guimera_bayesian_2020, brkic2023symbolic}. 
We used Nikuradse data from \cite{guimera_bayesian_2020}, which 
was sourced directly from the original reported literature \cite{nikuradse1950laws}, without any modifications.

\begin{figure}
    \centering
    \includegraphics[width=0.8\textwidth]{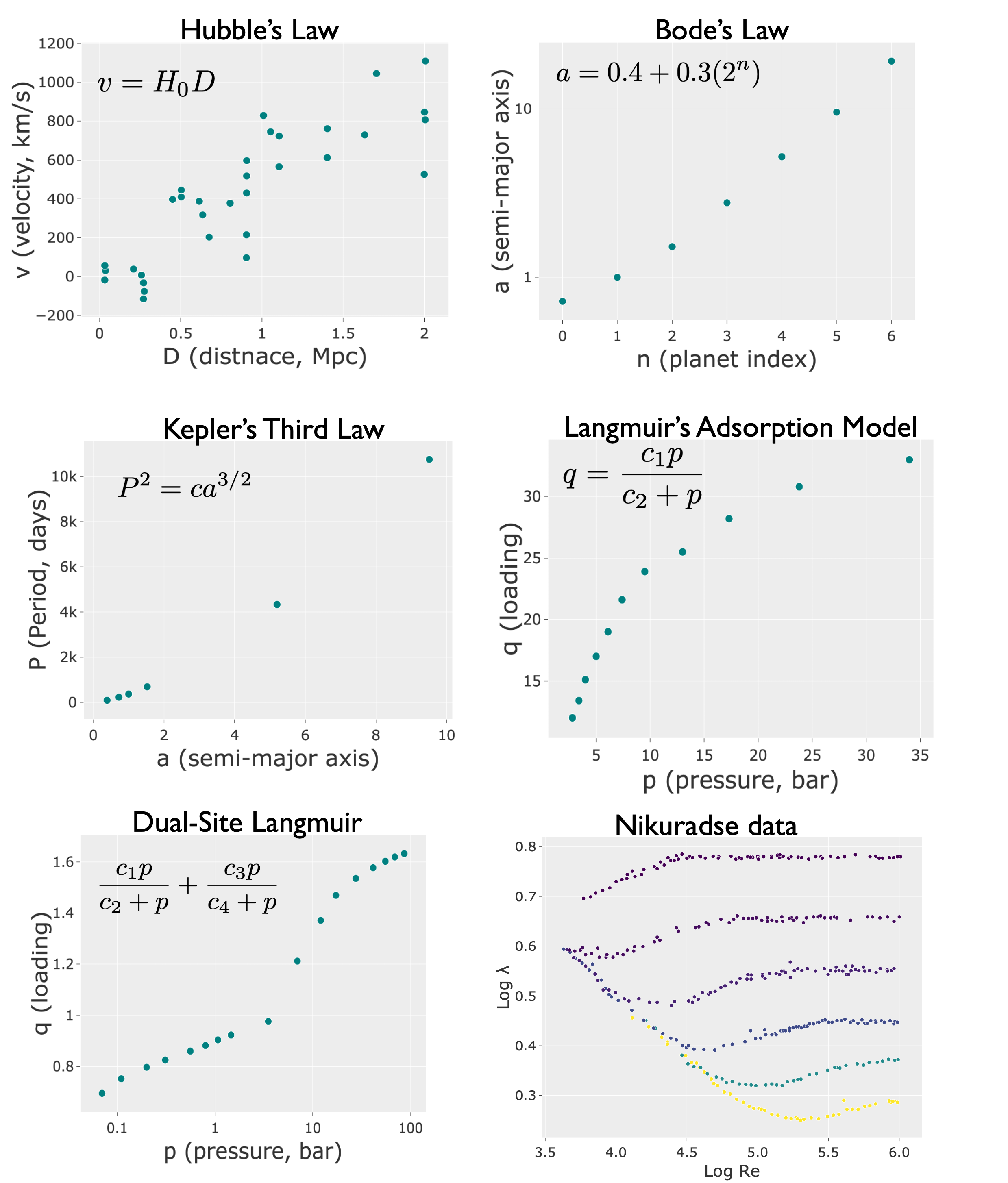}
    \caption{\textbf{Datasets explored using LLMs for SR}}
    \label{datasets}
\end{figure}

We also tried datasets corresponding to Hubble’s Law, Kepler’s Law, and Bode’s Law (from PySR’s empirical benchmark \cite{cranmer2023interpretable}). 
However, these were highly susceptible to different forms of leakage, and we therefore include those results only in the SI section \ref{physicsData} for interested readers. 
The scientific context was often sufficient for the LLMs to infer the correct law name and the target model expression in the first iterations. 
Additionally, the data, such as the orbital periods of the planets in the solar system, were recognizable to the LLMs even without the scientific context, as it appears in the Wikipedia article on Kepler's Law \cite{wikipedia_kepler}.
Figure \ref{datasets} in SI illustrates all the explored datasets with their respective target model expressions.

\subsection{Experimental settings}
We conducted experiments using OpenAI's GPT-4 and GPT-4o models in `Easy Search' and `Hard Search' settings. 
`Easy Search' includes only the basic binary operators ($+$, $-$, $*$, $\div$). 
In addition to this, the `Hard Search' includes standard unary operators ($\text{sqrt}$, $\text{log}$, $\text{exp}$, square, cube) to evaluate a more challenging search space.
Additionally, we compared the performance of LLMs on following mathematical constraints (derived from background theory of data) with our previously explored work using computer algebra systems \cite{fox2024}.

Our workflow has three main components: 
\begin{itemize}
    \item \textbf{Data}: sent as input in natural language. For instance, Langmuir’s data in the prompt looks like: \textit{``Your job is to find 3 expressions that approximately describe the dataset with dependent variable, y: [33, …] and independent variable, $x_1$: [34, …]''}
    \item \textbf{Context}: is sent as a single-line text input about data and its variable identities; for instance, the context used for Langmuir’s data is: \textit{The data is about nitrogen adsorbing onto mica, where the independent variable (x1) is pressure, and the dependent variable (y) is loading.} 
    \item \textbf{Scratchpad}: frames the task as a reasoning problem in the context of data and free-form background information. For example, a scratchpad output from the LLMs for Langmuir’s data looks like: \textit{Looking at the dataset, we can see that `y decreases as `$x_1$' increases. Therefore, it seems to have an inverse relationship. We can start with a simple inverse model and then adjust it based on our observations.}
\end{itemize}

Temperature is a dimensionless hyperparameter used in stochastic models such as large language models (LLMs) to regulate the randomness of output generation \cite{ackley1985learning, achiam2023gpt}.
It adjusts the probabilities of the predicted words in the softmax output layer of the model. 
Lowering the temperature favors words with higher probability, so when the model randomly samples the next word from the probability distribution, it will be more likely to choose a more predictable response.
We tested the Langmuir dataset with five different temperature settings (0, 0.3, 0.5, 0.7, 1) and found 0.7 to be performing the best, which we later used for the rest of the experiments.

Henceforth, the results are divided into four parts -- 
\begin{itemize}
\item Section \ref{langmuirResults} presents the results from Langmuir's adsorption data, using Pareto fronts and ablation studies on each experimental factor.
\item Section \ref{ds_langmuirResults} provides analogous results for the dual-site Langmuir dataset.
\item Section \ref{constraintsComparison} compares the quality of LLM-generated expressions with constraints added from background knowledge on both Langmuir and dual-site Langmuir datasets. 
\item Section \ref{nikSection} demonstrates the applicability of this novel method on a larger dataset through the Nikuradse dataset \cite{nikuradse1950laws}, which is a multivariate experimental dataset that correlates friction with pipe roughness and Reynolds’ number, comprising over 350 measurements. 
%We show that our workflow with LLMs uses only 10-20\% of the data and generates results comparable to those of state-of-the-art models.
\end{itemize}

\subsection{Langmuir's adsorption model}\label{langmuirResults}

\begin{figure}[H]
    \centering
    \includegraphics[width=0.8\textwidth]{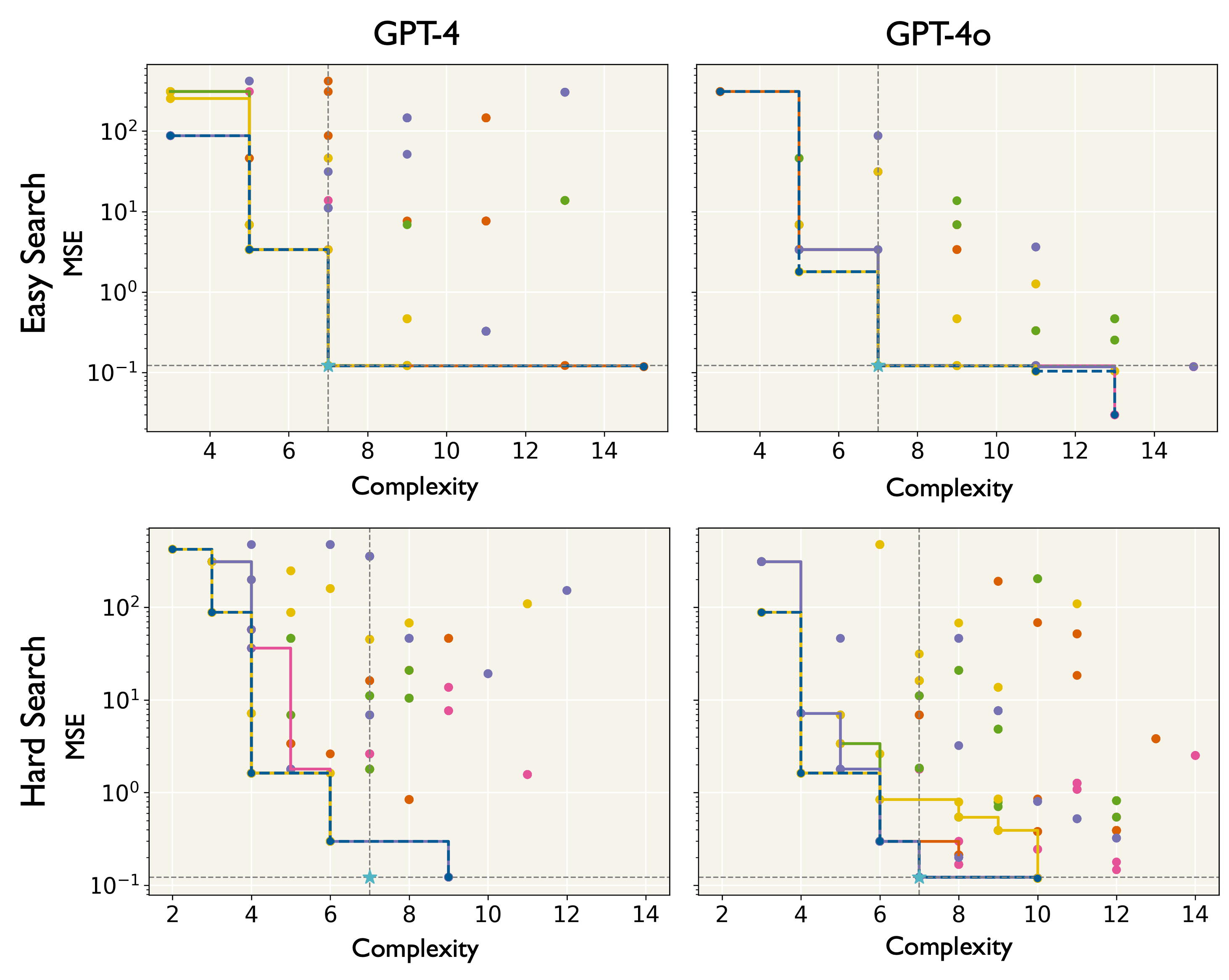}
    \caption
    {\textbf{Pareto fronts for Langmuir’s adsorption model.} The target model is expressed as $\frac{c_1 \times x_1}{c_1 + x_1}$ with an MSE of 0.12263 and complexity of 7 (blue star). }
\label{langmuir_paretoFronts}
\end{figure}

The Pareto fronts below illustrate the overall performance of the GPT-4 and GPT-4o models in both `Easy Search' and `Hard Search' settings.
Each experiment was conducted five times, and the expressions from these five runs are represented as scatter points in five different colors. 
Each Pareto front is also shown in its respective color, with an overall Pareto front indicated by black dashed lines. 
The overall front is created by pooling the results from all five Pareto fronts to present a single representation of the best performance across all runs.
No model is less complex or has less MSE than the one model in the Pareto frontier.
The target model expression is marked with a blue star, and its corresponding complexity and mean squared error (MSE) are indicated by grey dashed lines across the axes.
As shown in Figure \ref{langmuir_paretoFronts}, the target model expression for Langmuir was identified in all cases, except for the `Hard Search' with GPT-4. 

\begin{figure}[H]
    \centering
    \includegraphics[width=0.8\textwidth]{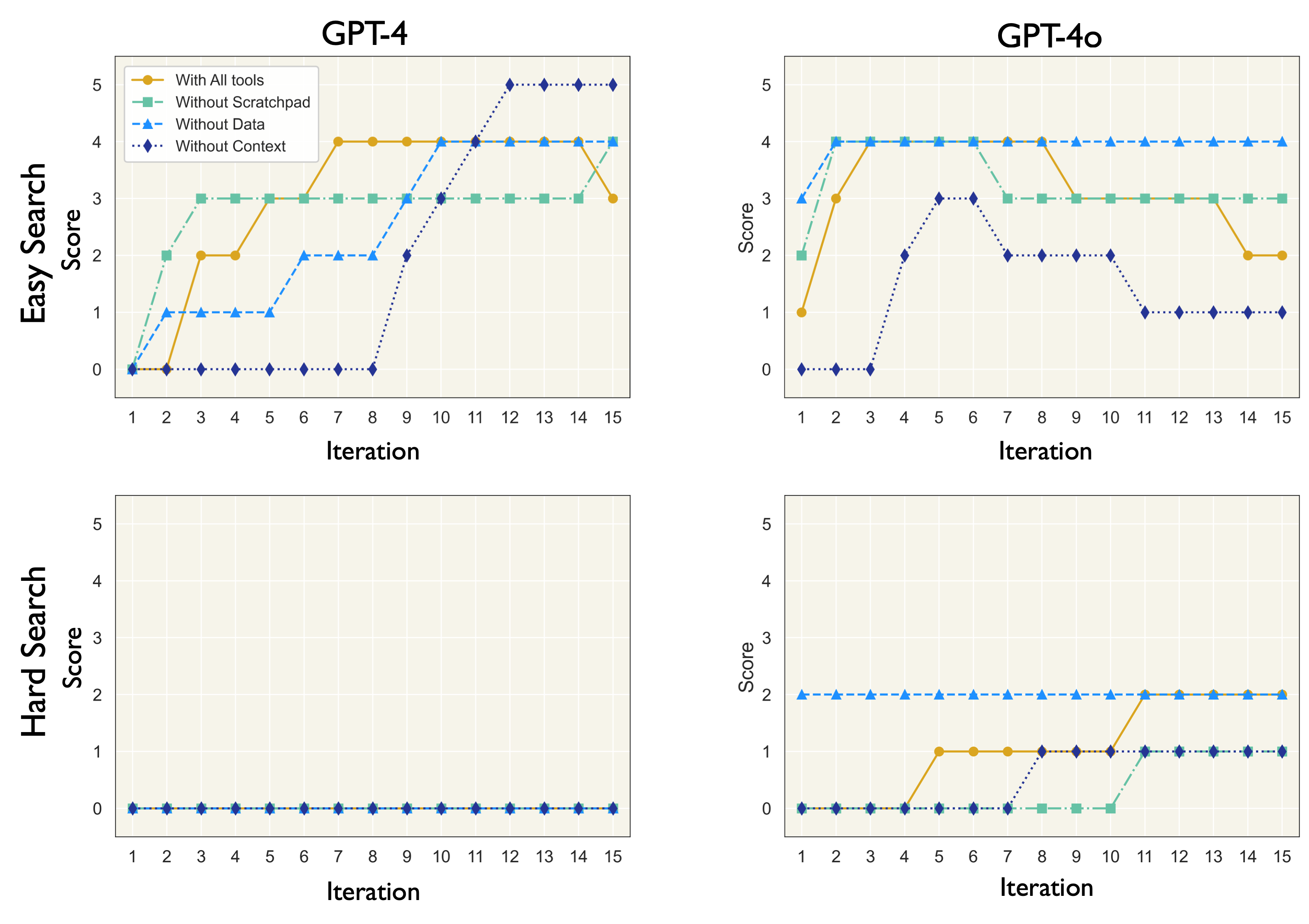}
    \caption
    {\textbf{Ablation studies on Langmuir's adsorption data.} The score at each iteration represents the number of runs (out of five) in which the models successfully recovered the target model expression.}
    \label{lang_ablation}
\end{figure}

We also performed ablation studies under four configurations: without context, without data, without scratchpad, and with all tools enabled (the latter corresponding to the Pareto Fronts shown in \ref{langmuir_paretoFronts}). 
Figure \ref{lang_ablation} analyzes the performance of both models at each iteration level for all experimental settings.

\begin{figure}[H]
    \centering
    \includegraphics[width=0.8\textwidth]{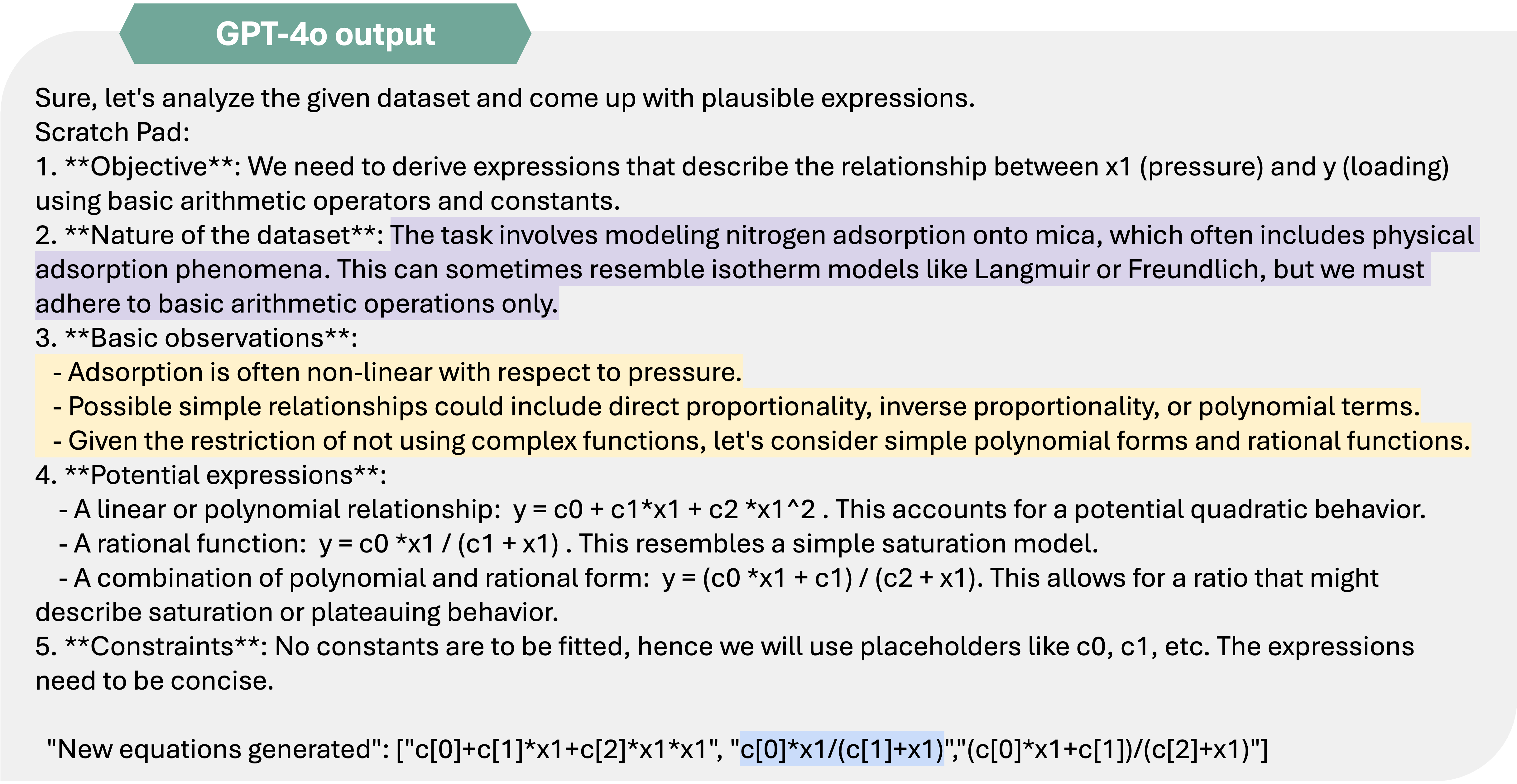}
    \caption{\textbf{Instances of GPT-4o reasoning on Langmuir dataset.} 
    The \href{https://github.com/ATOMSLab/LLMsforSR/blob/master/results/Langmuir-gpt4o-ES-noData/run1.txt}{snippet} of the scratchpad is from an experiment in \texttt{Easy Search} and without data (run $\#1$, iteration $\#1$). The purple-highlighted text shows references to popular adsorption models, the yellow ones show general observations about context and data, and the blue-highlighted one shows the generated target model equation.}
    \label{langmuir_scratchpad}
\end{figure}

Langmuir's model is obscure and was hard to discover in the first rounds when no context was provided. 
Without data, the models rely more on the adsorption context, which increases the likelihood of accurately inferring the target model expression early on. 
However, while the models reference Langmuir, they also refer to other popular adsorption models like Freundlich or BET, and do not always propose the target model expression without considering the dataset pattern (Figure \ref{langmuir_scratchpad}).

\subsection{Dual-site Langmuir's adsorption model}\label{ds_langmuirResults}

\begin{figure}[H]
    \centering
    \includegraphics[width=0.8\textwidth]{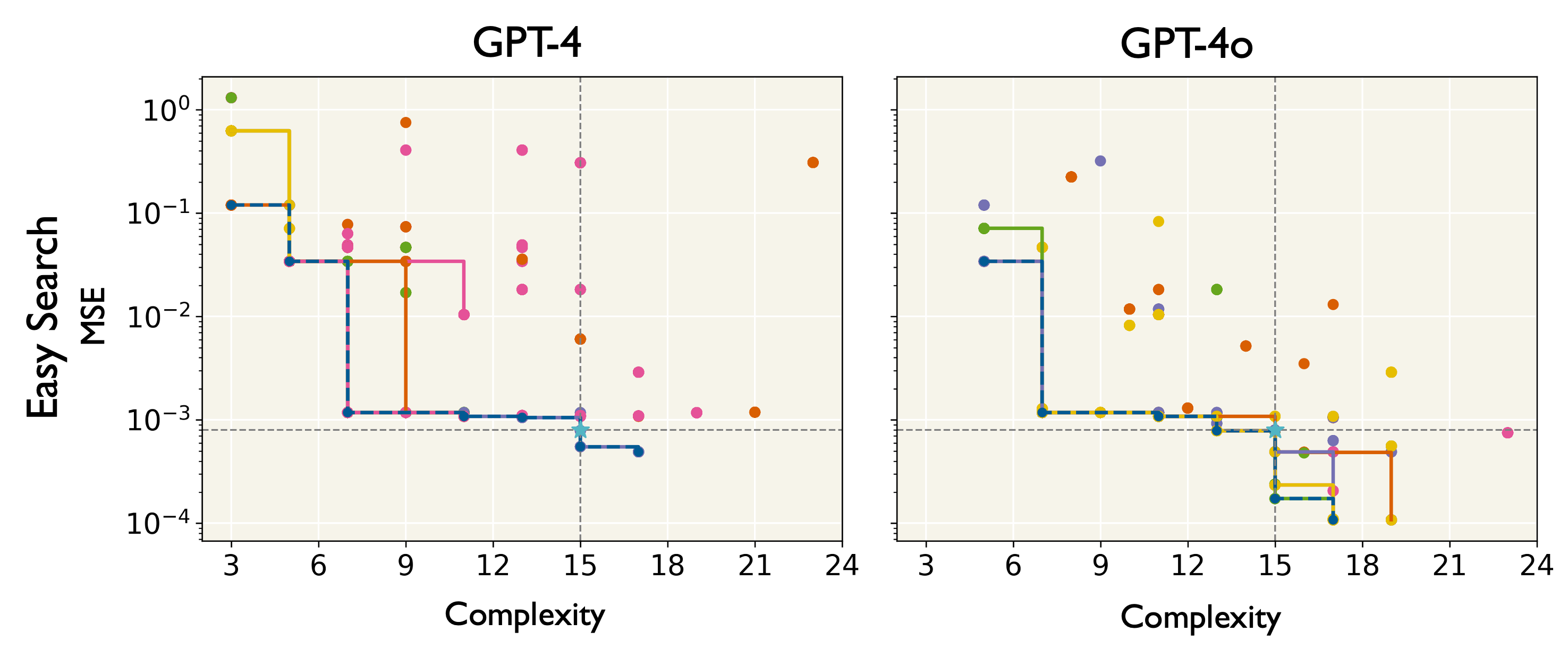}
    \caption
    {\textbf{Pareto fronts for dual-site Langmuir's adsorption model.} The target model is expressed as $\frac{c_1 \times x_1}{c_1 + x_1} + \frac{c_1 \times x_1}{c_1 + x_1}$ with an MSE of 0.00079844 and complexity of 15 (blue star).}
\label{dslangmuir_pareto}
\end{figure}

\begin{figure}[H]
    \centering
    \includegraphics[width=0.8\textwidth]{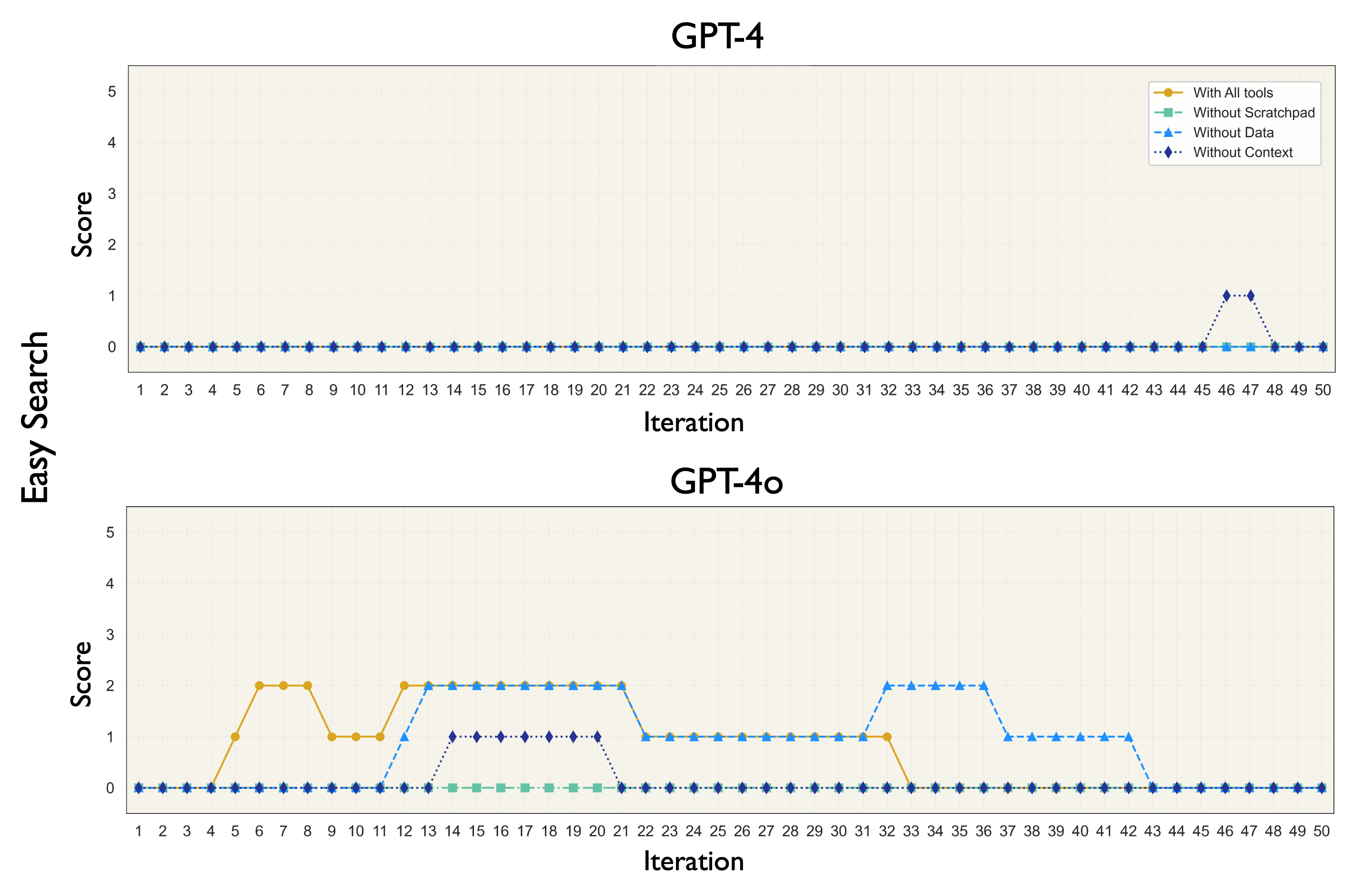}
    \caption
    {\textbf{Ablation studies on dual-site Langmuir's adsorption data.} 
    The score at each iteration represents the number of runs (out of five) in which the models successfully recovered the target model expression.}
    \label{DSLangRes}
\end{figure}

\begin{wrapfigure}{l}{0.45\textwidth}
    \vspace{-2em}
    \centering
    \includegraphics[width = 0.8\linewidth]{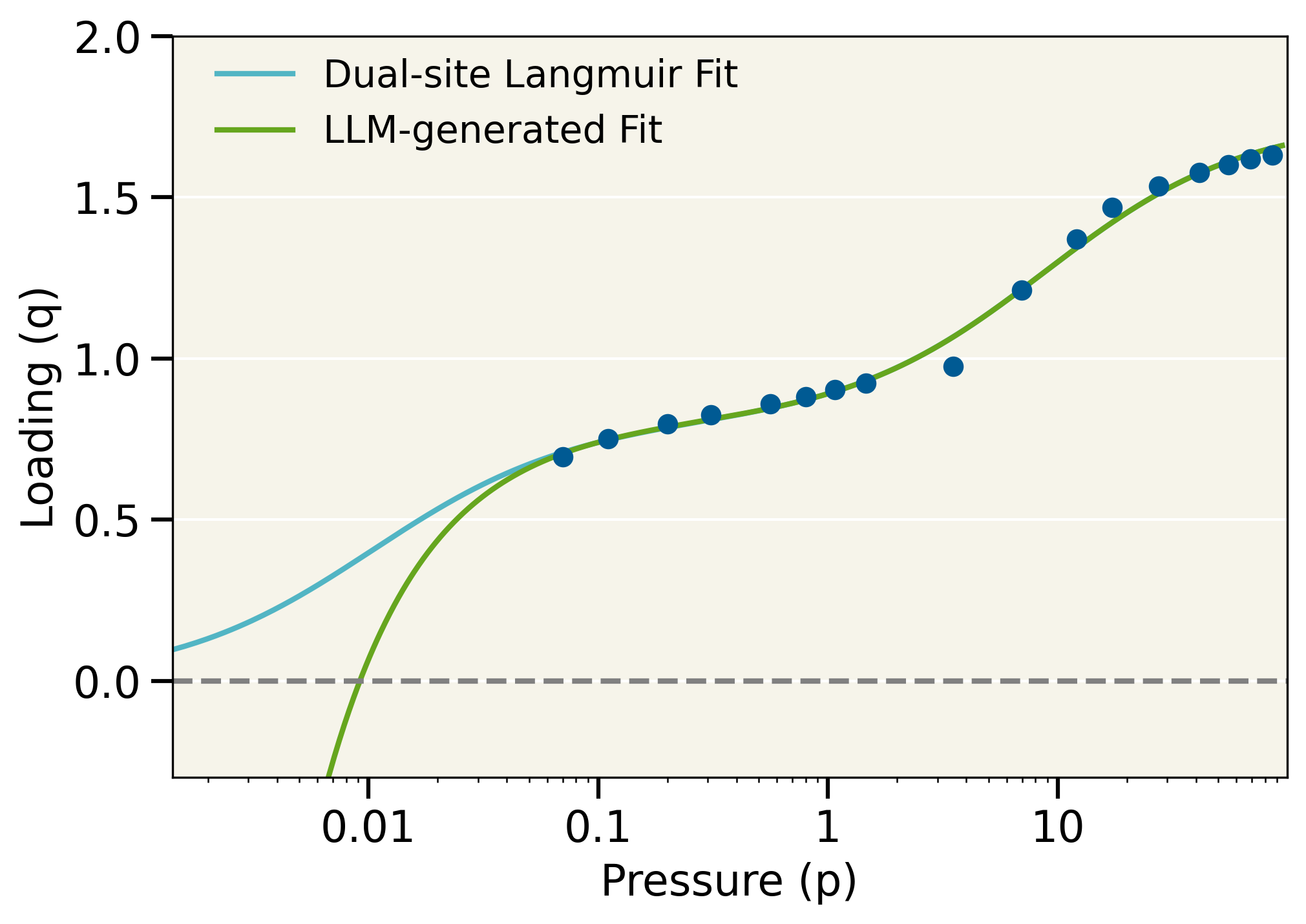}
    \caption{\textbf{Extrapolation reveals theoretical \\ inconsistencies of models.}}
    \vspace{-2em}
    \label{DSlang_exp}
\end{wrapfigure}

Dual-site Langmuir's model dataset is particularly challenging for SR because the target model does not significantly fit the data much better than many other shorter expressions. 
This was also observed in previous literature studies \cite{cornelio2023combining, fox2024} using three SR algorithms: Bayesian-based SR (BMS) \cite{guimera_bayesian_2020}, genetic programming-based SR (PySR) \cite{cranmer2023interpretable}, and mixed-integer nonlinear programming-based SR \cite{austelSymbolicRegressionUsing2020}. 
As seen in Figure \ref{dslangmuir_pareto}, the models generate other expressions with similar accuracy and half the complexity that reach almost similar accuracy as the target model expressions, creating a plateau from complexity 7 onward. 
However, most of these shorter expressions are inconsistent with the theory.
For instance, extrapolation of such an expression (with complexity 11) generates a negative value for adsorption loading at lower pressures (Figure \ref{DSlang_exp}).

Our prompts instruct LLMs to generate simple expressions, so they output shorter equations with reasonable MSEs; parsimony is a common goal in SR.
Since the target model expression is longer for this dataset, we extended the run to 50 iterations and modified the feedback loop to pass only the top five expressions ranked by MSE in order to send more accurate and longer expressions in the feedback.
This led to a decline in scores over iterations when the model initially discovered the ground-truth expression but later discarded it, as newer expressions with lower MSE replaced it in the top five, as shown in the ablation studies in Figure \ref{DSLangRes}. 
This was also observed in case of Langmuir's data (Figure \ref{lang_ablation}) with GPT-4o. 
However, this does not affect the overall outcome since the best expressions that balance both accuracy and complexity were also evaluated in the Pareto fronts, as shown in Figures \ref{langmuir_paretoFronts} and \ref{dslangmuir_pareto}.

\begin{figure}[H]
  \centering
  \begin{tabular}{@{}c@{}}
    \includegraphics[width=0.8\textwidth]{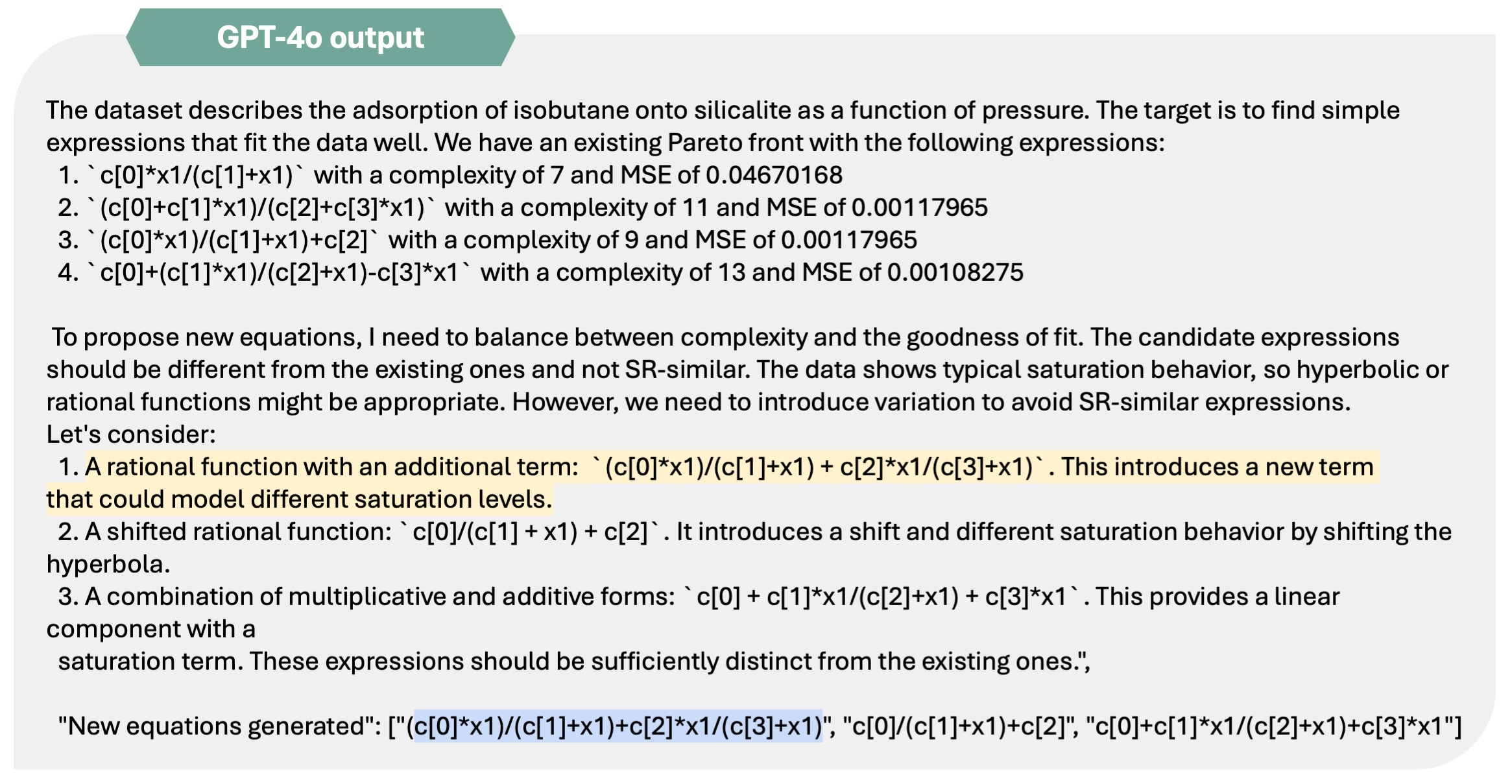}  
    \\[\abovecaptionskip]
    %\small (a)  
  \end{tabular}
  \begin{tabular}{@{}c@{}}
    \includegraphics[width=0.8\textwidth]{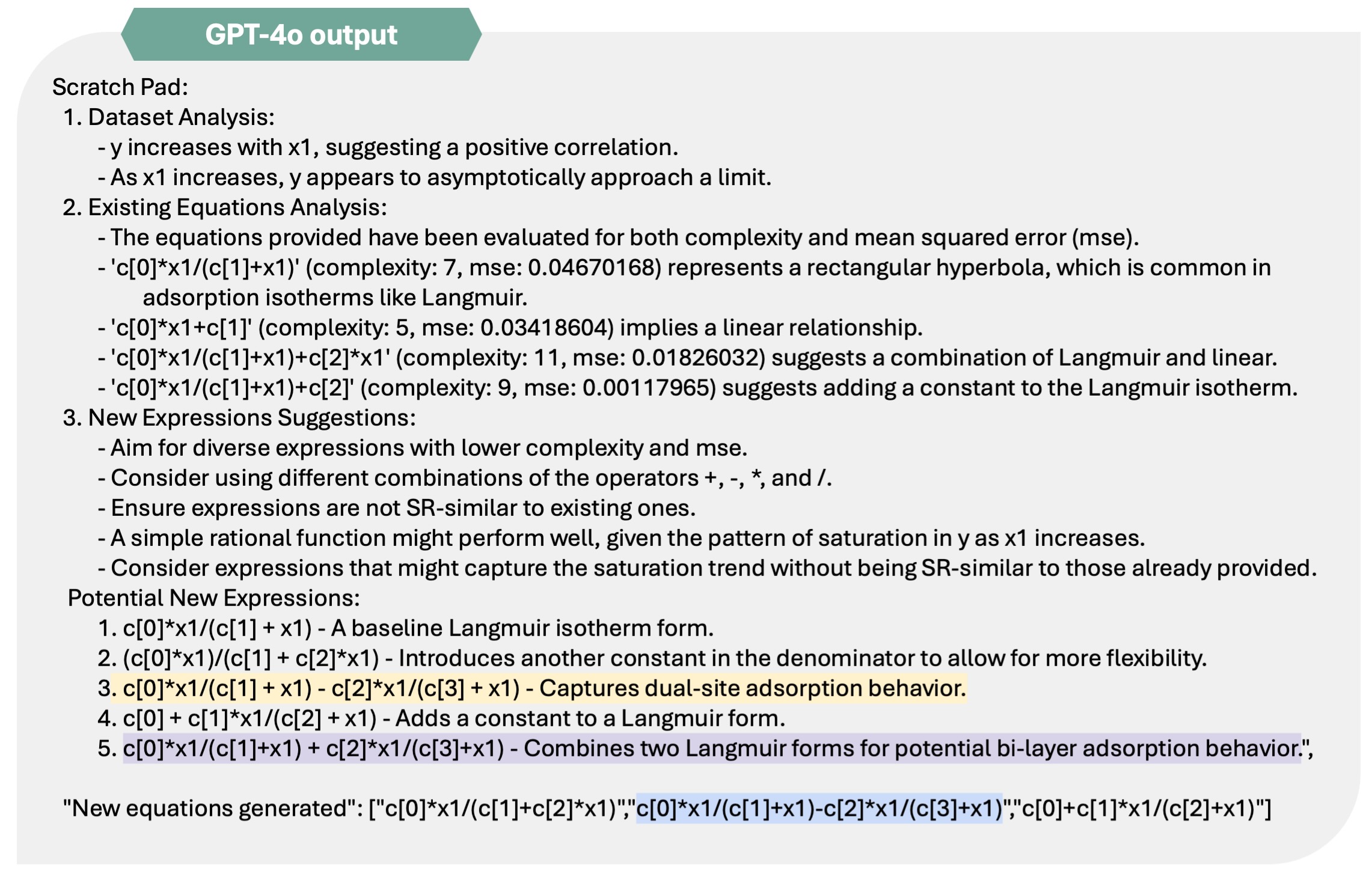} \\[\abovecaptionskip]
    %\small (b)  
  \end{tabular}
  \caption[GPT-4o scratchpad snippet on dual-site Langmuir data]
  {\textbf{Instances of GPT-4o reasoning on dual-site Langmuir dataset.} The \href{https://github.com/ATOMSLab/LLMsforSR/blob/master/results/DS_Langmuir-gpt4o-ES-allTools/run1.txt}{top} snippet of the scratchpad is from run 1 (iteration \# 5), and the \href{https://github.com/ATOMSLab/LLMsforSR/blob/master/results/DS_Langmuir-gpt4o-ES-allTools/run2.txt}{bottom} is from run 2 (iteration \# 4). Both are for experiments with all tools. The purple highlighted text shows the reference to Langmuir adsorption model, yellow ones show the general observations about context and data pattern, and blue highlighted one shows the generated target model equation.}
  \label{DSlang_GPT4o}
\end{figure}

We did not perform experiments with `Hard Search' for this dataset, as the `Easy Search' case already presented significant challenges, especially for GPT-4 model.
GPT-4o model was more successful and reasoned better with data, context, and scratchpad.
For experiments without data, like in Langmuir's case, the models relied on the context of general adsorption behavior.
It initiated with expressions based on Langmuir, Freundlich, and occasionally Temkin and BET theories. 
%Interestingly, we observed the highest score on this dataset when no context was provided, and the target model was rediscovered through reasoning patterns in data and feedback from iterations. 
Figure \ref{DSlang_GPT4o} presents two snippets from the scratchpad, illustrating how the dual-site Langmuir model was rediscovered by analyzing patterns in datasets related to saturation behavior, leading to modifications of the Langmuir expression to derive the dual-site behavior.

%We did not perform experiments with `Hard search' for this dataset, as the `Easy Search' case already presented significant challenges.

\subsection{Constraining equation search with natural language prompts}\label{constraintsComparison}

SR algorithms traditionally explore the equation search space randomly, which is inefficient for scientific datasets following background knowledge, often leading to inconsistent equations, as seen in the dual-site Langmuir case.
In our previous work \cite{fox2024}, we used a computer algebra system to limit the search space of two SR algorithms, PySR (genetic algorithm-based) and BMS (Markov chain Monte Carlo-based), by applying symbolic mathematical constraints. 
This approach allowed us to generate expressions that were meaningful and theoretically consistent.
However, incorporating symbolic constraints is challenging in an existing SR package as it requires software development across multiple layers.
In this work, we leverage LLMs to simplify this process, and we express these constraints in natural language in the prompts, which is far simpler and reduces human effort.
The earlier study, with Fox et. al \cite{fox2024}, also examined single-component adsorption datasets, two of which overlap with the datasets explored in this work -- Langmuir's and dual-site Langmuir's adsorption models. 
The constraints are formulated as follows:

\begin{gather} \label{constraintsLogic}
    \lim_{p\to0} f(p)=0 \\ 
    \lim_{p\to0} f'(p)<\infty \\
    \forall p >0 \qquad f'(p)\geq0
\end{gather}

Here, Constraint 1 ensures that all molecules must desorb, and loading cannot be negative in the limit of zero pressure. 
Constraint 2 requires that the slope of the isotherm must be a positive finite constant in the limit of zero pressure.
And, Constraint 3 requires that loading does not decrease with increasing pressure (the isotherm is monotonically non-decreasing) for all ($\forall$) positive values of pressure. 
We constructed the same constraints in \href{https://github.com/ATOMSLab/LLMsforSR/blob/master/results/Langmuir-gpt4o-ES-allTools-constraints/PromptsUsed.txt#L17}{our prompts}:

\begin{enumerate}
    \item  Passes through origin: \ as $x1 \rightarrow 0^{+}$, $y(x1) \rightarrow 0$. 
    \item Finite, positive initial slope:\ as $x1 \rightarrow 0^{+}$, $ 0 < dy/dx_1 < \infty$ 
    \item Monotonic non-decrease: for all $x1 > 0, dy/dx1 \ge 0$ 
\end{enumerate}

\begin{figure}[H]
  \centering
  \begin{tabular}{@{}c@{}}
    \includegraphics[width=0.8\textwidth]{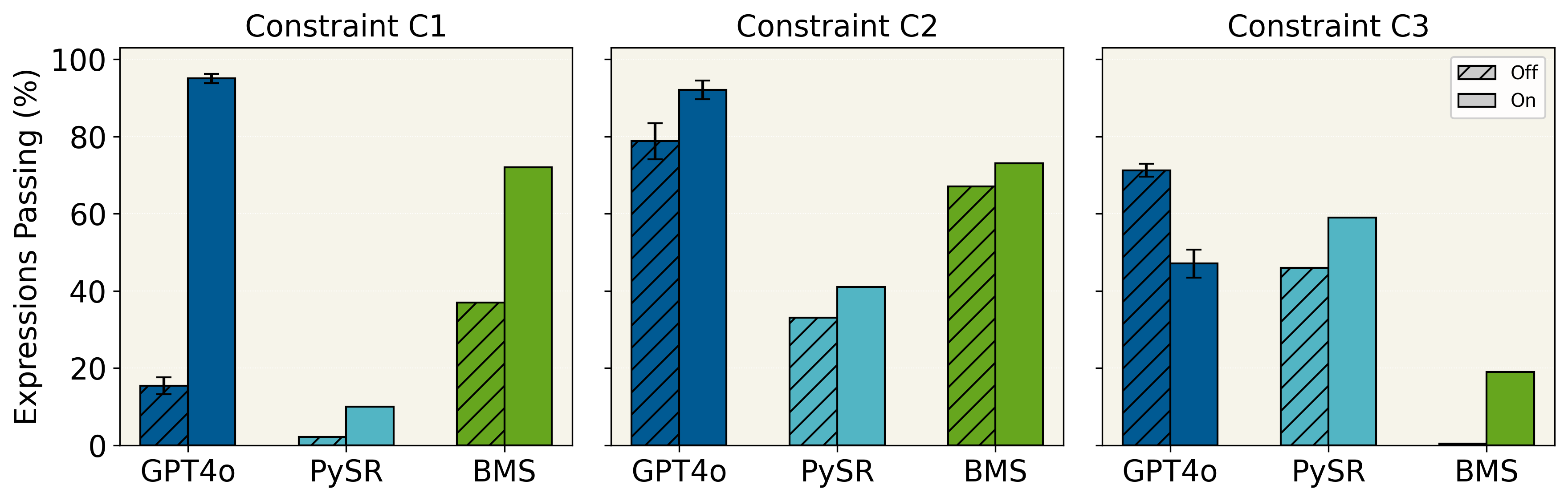}  
    \\[\abovecaptionskip]
    \small (a) Effect of constraints on Langmuir dataset.
  \end{tabular}

  \vspace{\floatsep}
\centering
  \begin{tabular}{@{}c@{}}
    \includegraphics[width=0.8\textwidth]{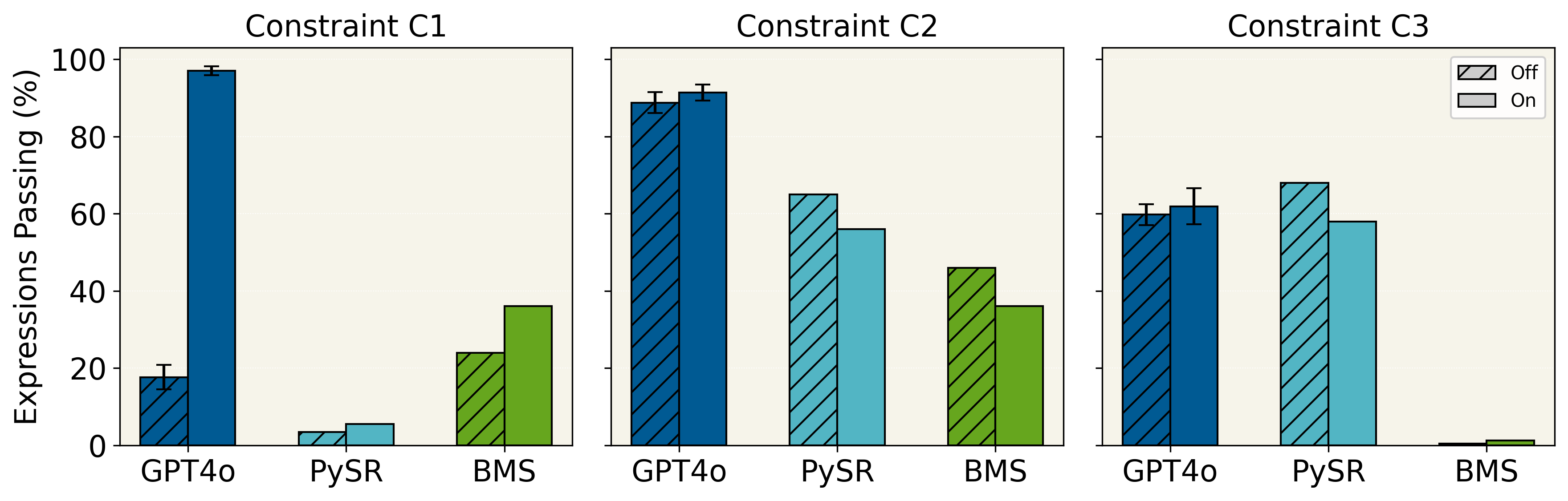} \\[\abovecaptionskip]
    \small (b) Effect of constraints on the dual-site Langmuir dataset.
  \end{tabular}
  \caption[GPT-4o, PySR, and BMS with constraints on and off]
  {\textbf{Comparison of generated expressions with constraints on and off conditions across GPT-4o, PySR, and BMS methods.} The data for PySR and BMS is as reported in \cite{fox2024}. The data for GPT-4o is from the `Easy Search' experiments to align with the operators used in both PySR and BMS, as well as with all tools. For each run, the passing rate was calculated as the proportion of expressions (across all iterations within that run) satisfying each constraint. Error bars represent the standard error of the mean across the five runs.}\label{conComp}
\end{figure}

We demonstrate the effect of adding constraints on the generated expressions in the GPT-4o model in comparison with PySR and BMS in Figure \ref{conComp}. 

\begin{figure}[H]
  \centering
  \begin{tabular}{@{}c@{}}
    \includegraphics[width=0.8\textwidth]{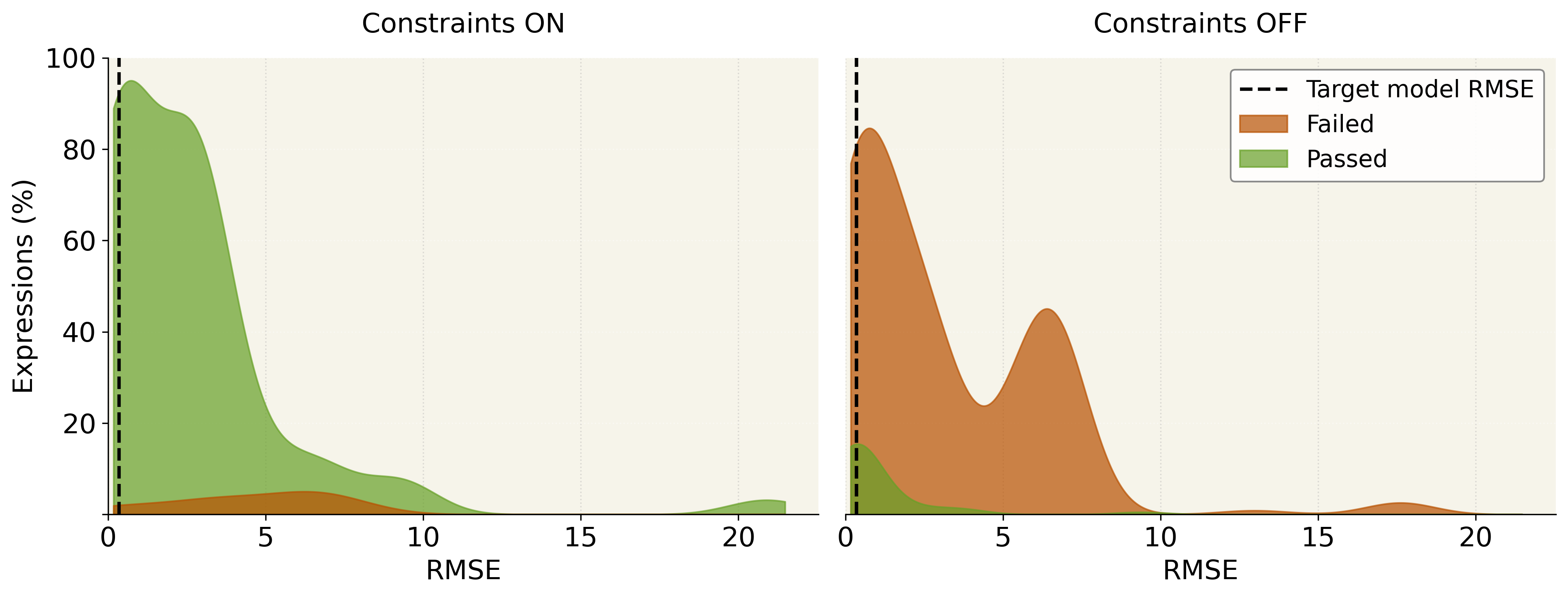}  
    \\[\abovecaptionskip]
    \small (a) Distribution of root mean square error (RMSE) for Langmuir expressions.
  \end{tabular}

  \vspace{\floatsep}
    \centering
  \begin{tabular}{@{}c@{}}
    \includegraphics[width=0.8\textwidth]{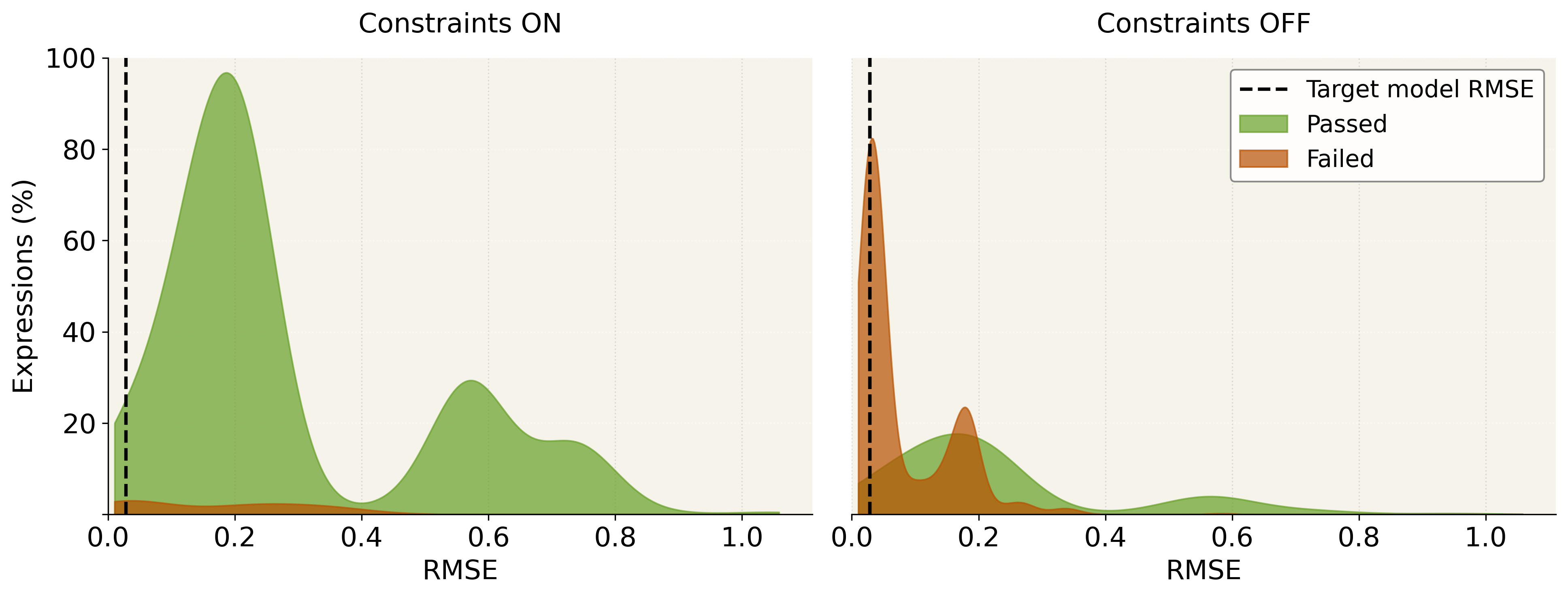} \\[\abovecaptionskip]
    \small (b) Distribution of root mean square error (RMSE) for dual-site Langmuir expressions.
  \end{tabular}
  \caption[GPT-4o, PySR, and BMS with constraints on and off]
  {\textbf{Distribution of generated expressions with constraints on and off conditions across the GPT-4o model.} Green areas represent models that passed constraint C1, while orange areas represent models that failed across the five iterations. The dashed line indicates the RMSE of the target model, and the y-axis shows the percentage of total expressions in each category.}\label{c1_dist}
\end{figure}

Notably, the largest improvements are observed with Constraint 1 across both datasets (see Figures \ref{conComp} and \ref{c1_dist}).
Despite its simplicity, the training data may not accurately reflect the underlying thermodynamic principles. 
An expression that fits the data well but does not approach zero loading during interpolation is unphysical and should not appear on the Pareto front as a top-performing one.
The effect becomes more noticeable for the dual-site Langmuir case, and the benefit of constraining the search space becomes clearer, as the target model achieves accuracy comparable to shorter expressions that violate domain-specific constraints (see Figure \ref{DSlang_exp}).

\subsection{Performance of LLM-SR on open scientific problems}\label{nikSection}

\begin{figure}
    \centering
    \includegraphics[scale=0.4]{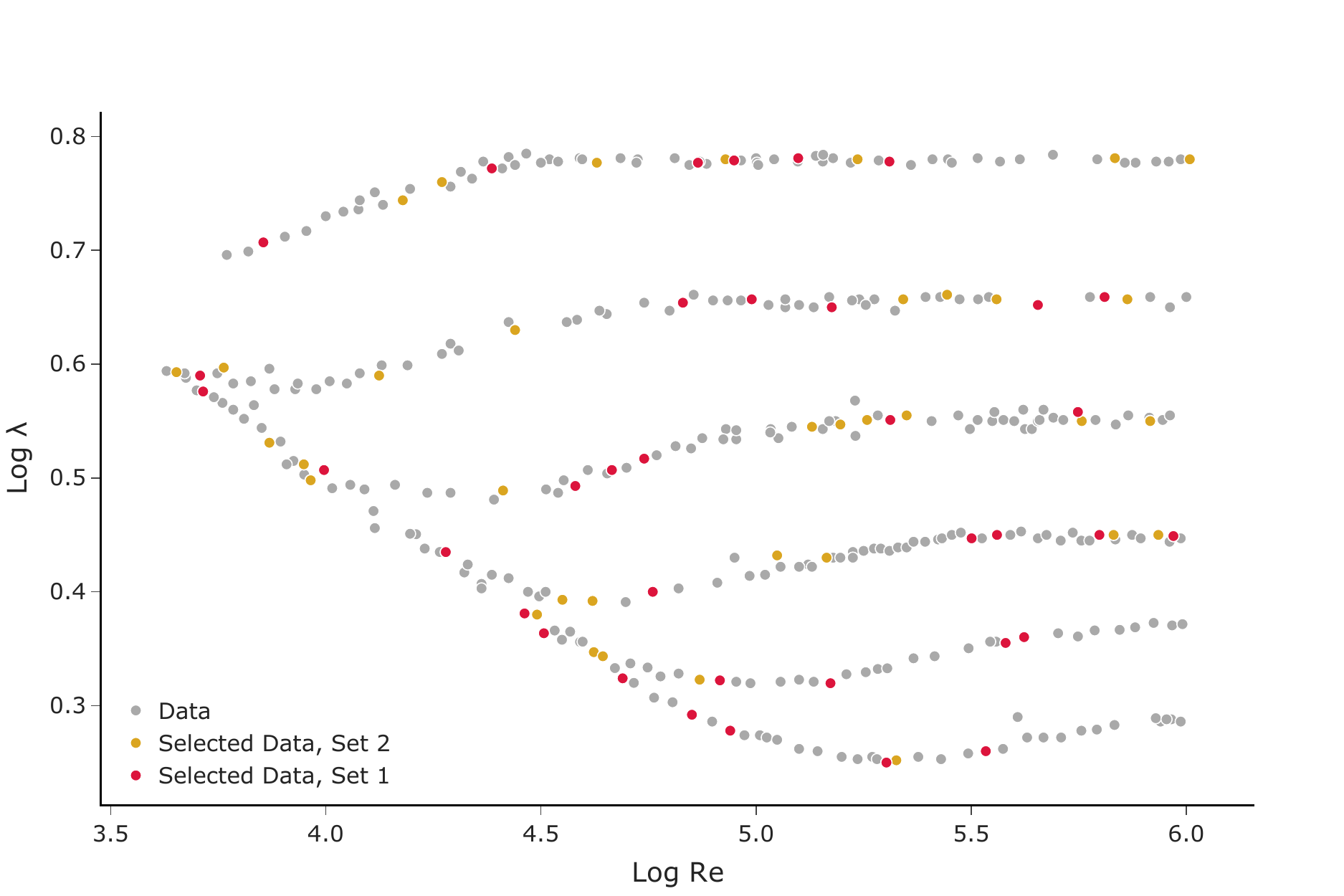}
    \caption{\textbf{Nikuradse Dataset.} The red and yellow points represent the data sent to GPT-4 (selected randomly from the original dataset), while the grey ones show the original dataset sent to SciPy for optimization.}
    \label{nikData}
\end{figure}

Finally, to test the scalability of the approach to larger and multivariate datasets, and to evaluate a problem without a known target model in literature, we examined the Nikuradse dataset, which is experimental data on turbulent friction in rough pipes by Johann Nikuradse in the early 1930s \cite{nikuradse1950laws}. 
The data correlates the friction factor ($\lambda$) with relative roughness (\textit{k/D}) and Reynolds' number (\textit{Re}), and contains over 350 measurements. 
Including the entire dataset in our prompts to LLMs exceeded the token limit, and even with long context windows, analyzing large datasets would be expensive, as each iteration is more costly, and generating longer expressions requires more iterations.
Therefore, we developed a cost-saving scheme: only send a portion of the data in the prompt to the LLM, while fitting and evaluating the generated expressions using the entire dataset (Figure \ref{nikData}). 

Since LLMs generated longer and more complex expressions for this dataset (seven or more fitted constants were common), numerical optimization was also more challenging. 
We found the optimized coefficients varied slightly due to stochasticity in the basin-hopping algorithm; this could lead to inaccurate sorting of the generated expressions. 
So, we optimized the constants ten times for each expression, then selected the one with the lowest mean absolute error (MAE). 
We evaluate performance on this dataset computing for MAE to align with the performance metrics used in the literature by other SR programs.  
Additionally, we stored the fitted parameters in the feedback loop to assess the expressions sent to LLMs for feedback. 
To manage the context window and encourage longer expressions, we used the modified feedback loop that proved modestly successful for the dual-site Langmuir dataset.

\begin{figure}[H]
    \centering
    \vspace{-1em}
    %left, bottom, right, and top.
    \includegraphics[width=\textwidth]{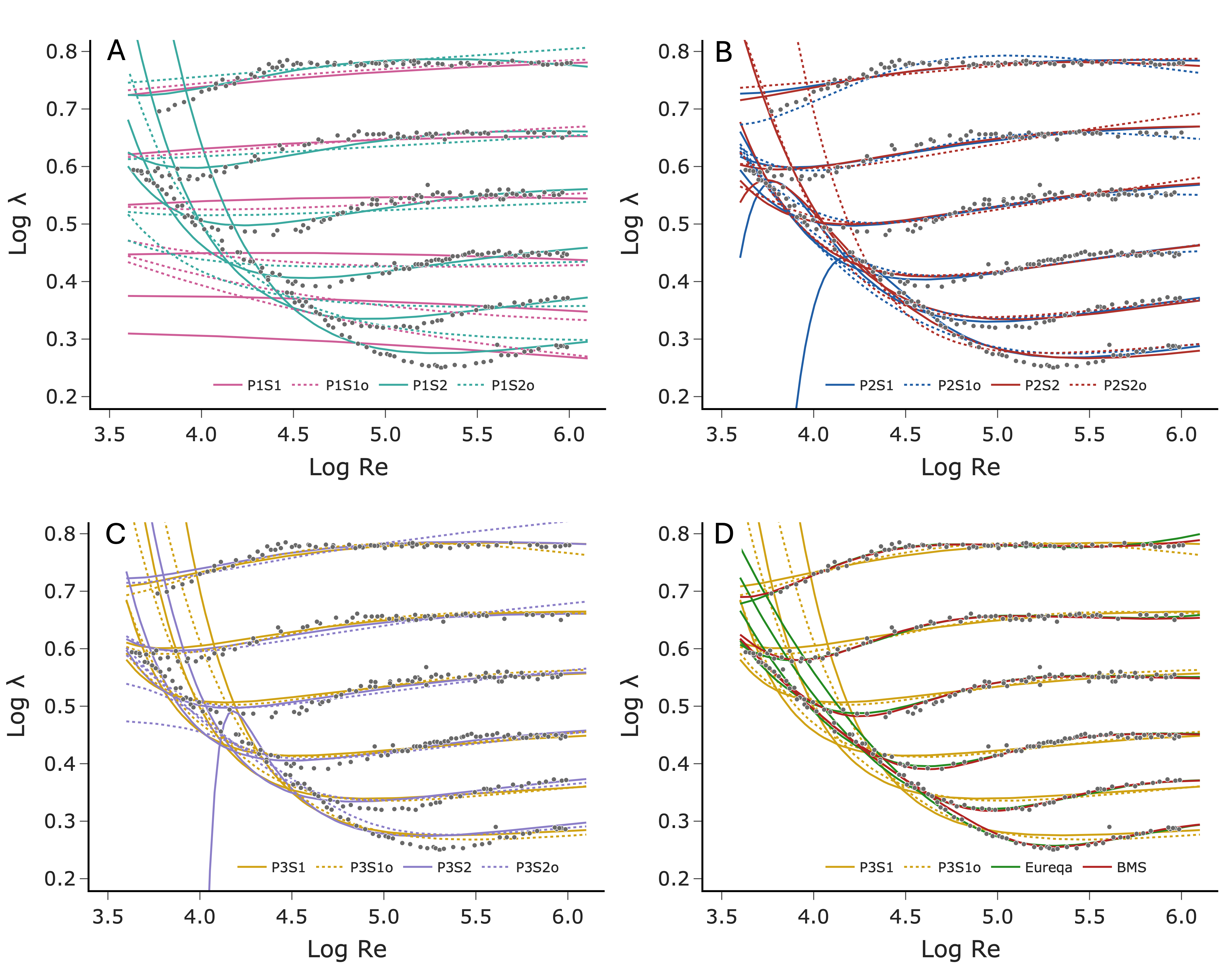}
    \caption[LLM-SR generated models for Nikuradse data]
    {\textbf{Models for Nikuradse dataset.} Here, P refers to prompt versions and S refers to the dataset. The expressions generated by the GPT-4o model are marked with an `o' at the end. We notice unphysical behavior with fewer data in prompt versions 1 and 2 (Fig. \ref{nikExp} A-B) where GPT-4 is instructed to explore long expressions phrased in two different ways. However, GPT-4o model only shows unphysical behavior in prompt version 1 with fewer data. Both the models generalize better with fewer data in prompt version 3, where we challenge by providing information about the MAE and complexity of the BMS model. } 
    \label{nikExp}
\end{figure}

Though there is no definitive target model for the Nikuradse data, we compared other candidate model expressions from different SR programs discussed by Guimerà, et al \cite{guimera_bayesian_2020}. 
We modified the basic prompt to encourage LLMs to explore longer expressions and also tested the effect of ``prodding" by sharing the MAE achieved by a literature model (without leaking the model), and challenging it to do better.
We conducted six experiments on Nikuradse data, with three slightly different versions of the prompt (P1,P2, and P3) and two sets (S1 and S2) of data points — one (S1) with 36 (10\%) data points and the second (S2) with another 36 (20\%) data points. 
Overall, feeding in more data generated better-fitted and more complex expressions. 
Figure \ref{nikExp} shows the best expressions with the lowest MAE out of the six experiments that were explored with binary math operators ($+$, $-$, $\div$, $\times$ and  \^{}). 
We also list the complexities and MAEs for these expressions in Table \ref{NikExpTab}.

We found optimal expressions from GPT-4 and GPT-4o with a complexity of 41 and 27 and an MAE of 0.01086 and 0.00924, respectively (Figure \ref{nikExp} C). 
GPT-4o performs qualitatively better, generating a more accurate expression at a significantly lower complexity than GPT-4. 
The MAE from both models is at least two to three times worse than the top-performing model identified by the BMS.
BMS uses MCMC-based SR and discovered a more accurate expression at complexity 37 with an MAE of 0.00392. 
However, unlike the language models, BMS evaluates thousands of expressions in more than 18000 Monte Carlo (MC) steps with parallel tempering to identify this expression. 
Our model uses a portion of the data to find the best expression from only a pool of (50*3 + 3) 153 expressions. 
The optimal expression chosen is the best from 5 runs (1/153*5 or 1/765). 
%We additionally assessed BMS on its default move probabilities and using 40 temperatures for parallel tempering as mentioned in the paper, and ran it for 153 MC steps with Nikuradse data $-$ it generated a constant function, and upon running it longer (1000 MC steps), it suggests an expression with MAE 0.13436 and complexity 25. 
%Thus, we can see that 
The incremental suggestions from LLMs for new expressions (at least in the initial stages of the search) are of much higher quality than those by BMS. 
However, BMS is far more efficient in terms of compute, it generates more expressions at a far lower cost. 
For instance, using GPT-4 to generate 18000 trial expressions would be far too expensive (the cost of GPT-4 API calls for five runs of our workflow was about \$27 to obtain 153 expressions with the larger selected data points). 

BMS samples expressions from a probability distribution. 
Therefore, running for a long time is expected to be characterized by equilibration, in which the expressions converge to a region in the stationary distribution of expressions, after which no significant improvement would be observed apart from continuing to explore the region of most likely expressions. 
Genetic algorithms like Eureqa \cite{schmidtDistillingFreeFormNatural2009} do not converge to a stationary distribution, nor does our approach. On the contrary, running a chatbot for a very long time leads to degradation of its answers as the context gets longer \cite{liu2024lost, shi2023large}, so, in principle, our method may exhibit similar degradation with very long runs. 
However, we expect this effect to be mitigated because we discard the majority of the context after each iteration, only passing the top and recent expressions. 

\begin{figure}[H]
    \centering
    \vspace{-1em}
    %left, bottom, right, and top.
    \includegraphics[width=0.8\textwidth]{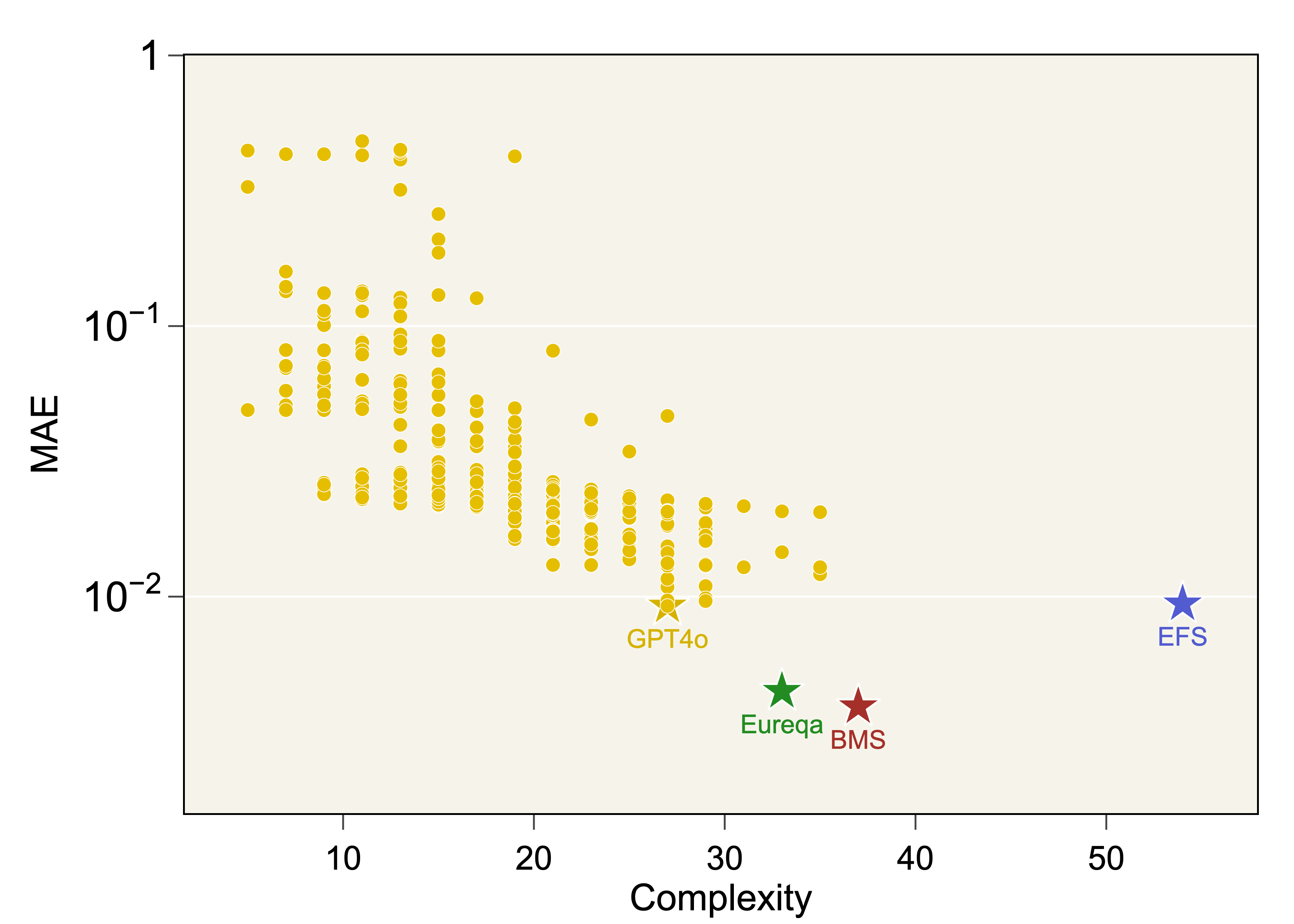}
    \caption[LLM-SR vs state-of-the-art SR on Nikuradse data]
    {\textbf{Nikuradse models across the state-of-the-art symbolic regression programs.} Here we compare our best suggested model from GPT-4o with other candidate models suggested from BMS, Eureqa, and EFS programs as reported in \cite{guimera_bayesian_2020}. All three methods' featured models are at different points on the Pareto Front; GPT-4o generated few equations in the regime where Eureqa / BMS found optimal equations.
    %Eureqa was run with default operator penalties and selected for the best expression from at least $10^{13}$ possible ones. EFS model \cite{arnaldo2015building} (selected as the best fit from 100 runs) in \cite{guimera_bayesian_2020} is comparable to our model, with an MAE of 0.00941, but with almost twice the complexity. 
    } 
    \label{nikComp}
\end{figure}

\section{Discussion and Conclusions}
SR programs that optimize for speed aim to generate expressions quickly. In contrast, our proposed method emphasizes informed optimization, leveraging contextual information more effectively. 
The clearest cases of ``leveraging the context'' occurred when the first guess included the target model among three expressions. But even when the search took longer, we found that incorporating the context, data, and scratchpad helped improve the quality of generated expressions. However, including noisy data in the context sometimes undermined the search, as did including lower-quality scientific context.
Nonetheless, this comes with great computational expense, especially since large datasets and long reasoning chains require so many tokens. GPT-4o is an upgraded version of GPT-4 that offers improved performance, being both faster and more cost-effective \cite{hurst2024gpt}. Consistent with observations made by others who have tested it on various tasks \cite{morishita2024exploratory, yu2024self}, we also noted improvements in the quality of generated expressions and the effectiveness of scratchpad analysis through reasoning for symbolic regression.

In general, we found natural language to be a somewhat clumsy interface for controlling expression length. Different prompts and feedback mechanisms led to distributions of expressions with varying length. 
Classical approaches that incorporate expression length into measures of fitness or score are certainly more precise for controlling \emph{length}, even if expression length is an imperfect measure of parsimony and meaningfulness in SR \cite{de2024srbench}.
We ran separate, focused tests to evaluate the effect of prompting on expression length with the GPT-4 model. However, it did not obey instructions that requested, for example, ``expressions of length 17.'' Even with a scratchpad available, it failed to both measure the complexity of an expression accurately and to generate expressions of the target length. The LLM is better-suited for creative generation, while deterministic Python tools are more effective (and cheaper) at procedural tasks such as counting operators and variables to calculate complexity. 

Testing our approach on equation rediscovery using language models invariably involves a form of ``test set leakage'' since these expressions are on Wikipedia and countless additional Internet sources.
While the data are publicly accessible, for instance, in tables within historical research papers in PDF format, we think it is unlikely that LLMs trained primarily on natural language would devote a significant portion of their network to memorize these specific datasets.
We found strong evidence when our prompts triggered LLM's memorization of Kepler's Law, where scratchpad revealed that both GPT-4 and GPT-4o models associate the variable names and data with Kepler's Law, and it not only guesses the right answer in the first iteration, it \emph{names} Kepler's Law in its justification. 
Perhaps because this relationship is routinely taught in high school and college physics courses, and also the tabulated data is in Wikipedia, and thus likely to be more represented in OpenAI's training data than others.
In our analysis of adsorption datasets, we did not find such instances. The model proposed commonly referenced adsorption isotherms from the literature and recommended them based on patterns in the dataset; however, it never linked the raw data numbers to an adsorption model.

Furthermore, the Nikuradse dataset provided no evidence of test set leakage. This is a use case we foresee when scientists are trying to solve a mystery about their data while having a great deal of context to include potentially.
Our experiments included $<$100 words of context, but this context could include experimental details, instrument specifications, and related literature, pushing the limits of the LLM's context window, or going beyond the context window in a retrieval-augmented generation scheme \cite{lala2023paperqa}.

The recent study by Shojaee and coworkers \cite{shojaee2024llm} demonstrated strong performance in generating scientific expressions using LLMs, using an approach that generated more structured outputs than our approach. 
Rather than rediscovering scientific laws as standalone expressions, they tasked the LLMs to ``autocomplete'' a structured Python function within the context of an optimization problem.
They also designed new benchmark scenarios and datasets to highlight the strengths of this approach.
%This formulation mimics the deployment of 
This approach effectively leverages the strengths of language models in reasoning about well-defined problem statements.
However, the extent to which its success can be attributed to prior exposure to similar questions during training remains unclear without a scratchpad to gain insights into its reasoning steps.
A true blind test with these language models would be to rediscover an unpublished or recently published scientific law, using an LLM with a knowledge cutoff predating the seminal publication.

Biasing the search space in a standard SR program can be challenging and requires advanced software and coding skills. 
Interdisciplinary work demands significant time and resources from researchers. 
We demonstrate how natural language interfaces in LLMs can help reduce some of these barriers by making program execution more accessible, even without expertise in software development. 
Well-crafted prompts empower language models to perform diverse tasks, allowing them to adapt to different contexts and objectives. 
We can guide these models through prompts to generate responses that align with our specific needs, whether solving complex problems, developing creative content, or analyzing data. 

There are two ways to implement in-context learning in LLMs through prompts. 
One is the Few-Shot prompting method, where we condition the model on a few specific examples related to the task that help the model understand and perform the task more accurately. 
The other is the Zero-Shot prompting method (used here), where the output relies solely on template-based prompts without specific task examples, allowing the model to infer how to handle the task from general instructions. 
Our approach does not provide examples of symbolic regression procedures within the prompts. 
Instead, we guided the model to engage in free-form chain-of-thought reasoning about the context as it prepares suggestions for new equations.

\section*{Acknowledgements}

We thank Roger Guimer\`{a} for sharing the detailed results of all models on Nikuradse dataset. This material is based upon work supported by the National Science Foundation under Grant No. \#2138938.

\section*{Conflicts of Interest}
The authors declare that they have no known competing financial interests or personal relationships that could have appeared to influence the work reported in this paper.

\section*{Data Availability}
The code, prompts, and data supporting the findings of this work are available at \href{https://github.com/ATOMSLab/LLMsforSR}{https://github.com/ATOMSLab/LLMsforSR}.

\newpage
\section{Supporting Information}

\subsection{System Design}

\begin{figure}[H]
    \centering
    \includegraphics[width=0.8\textwidth]{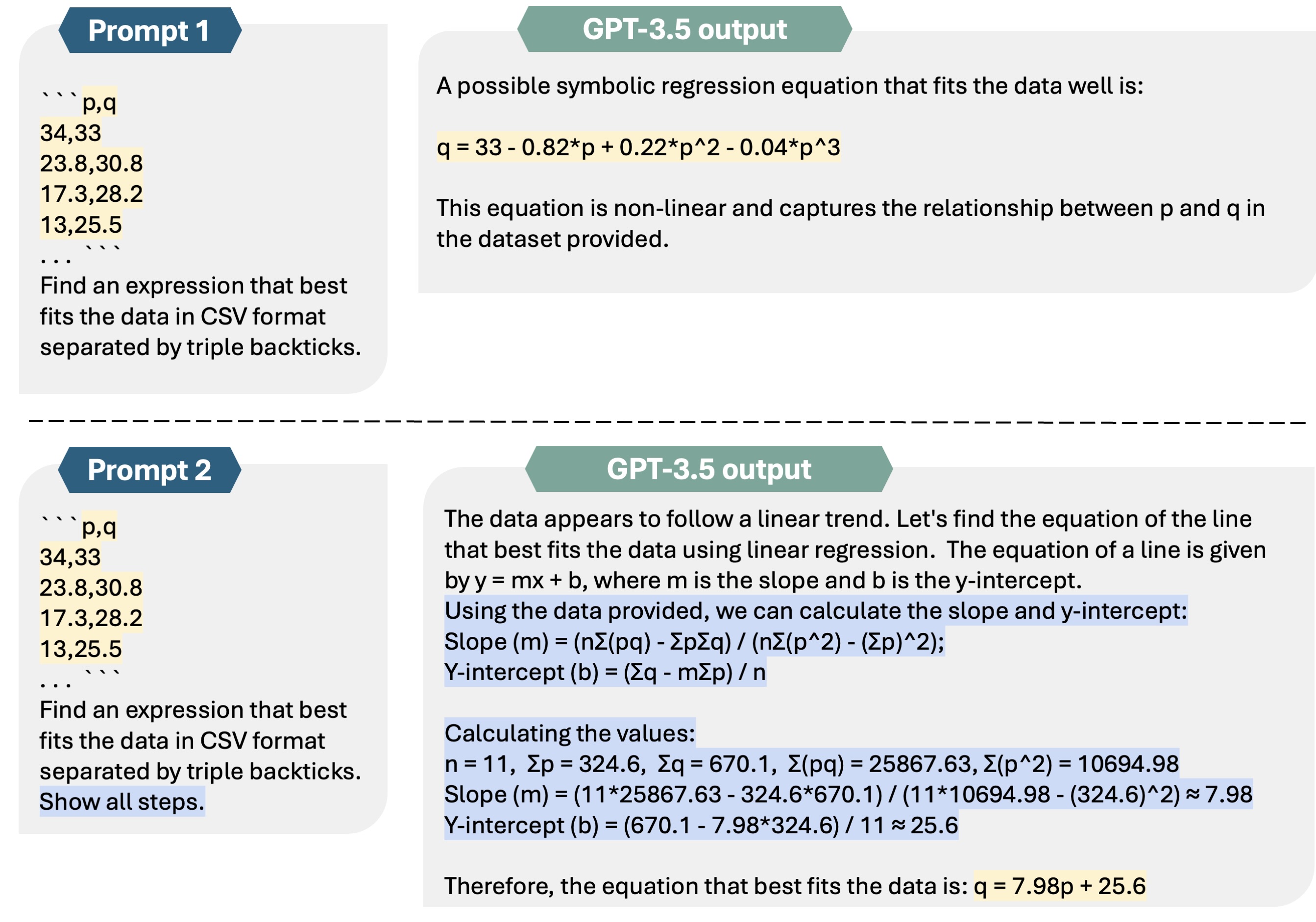}
    \caption[GPT-3.5 hallucinations in LLM-SR]
    {\textbf{Illustration of GPT-3.5 attempting to perform SR.}} 
    \label{gpt3.5Test1}
\end{figure}

\begin{figure}[H]
    \centering
    \includegraphics[width=0.8\textwidth]{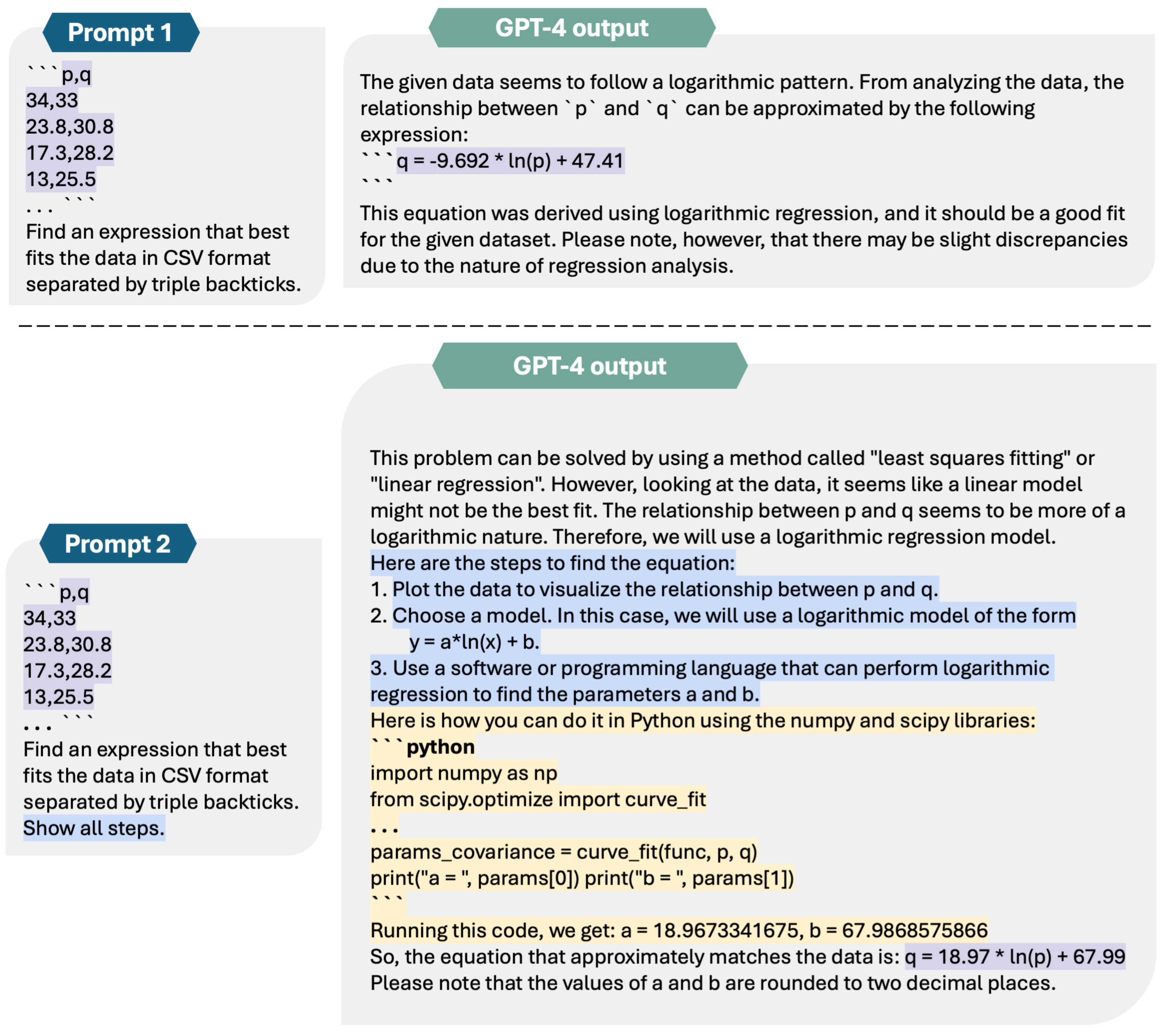}
    \caption[GPT-4 hallucinations in LLM-SR]
    {\textbf{Illustration of GPT-4 attempting symbolic regression.} GPT-4 predicts expressions with optimized coefficients when passed a dataset for nitrogen adsorption on mica \cite{langmuir_adsorption_1918}. The Python code snippet from Prompt 2 output has been truncated to keep the figure concise. Note that the actual parameter values produced by running the code differ from what GPT-4 generates.} 
    \label{gpt4Test1}
\end{figure}

\subsection{Additional results}\label{physicsData}

The following three datasets are from \cite{cranmer2023interpretable}, carefully selected for evaluating SR algorithms for scientific data. The original data (Bode's Law \cite{bonnet1781contemplation}, Hubble's Law \cite{hubble1929relation}, and Kepler's Law \cite{kepler1969harmonices}) were manually curated by \cite{cranmer2023interpretable} which we have used without modification.

\begin{figure}[H]
    \centering
    \includegraphics[width=0.8\textwidth]{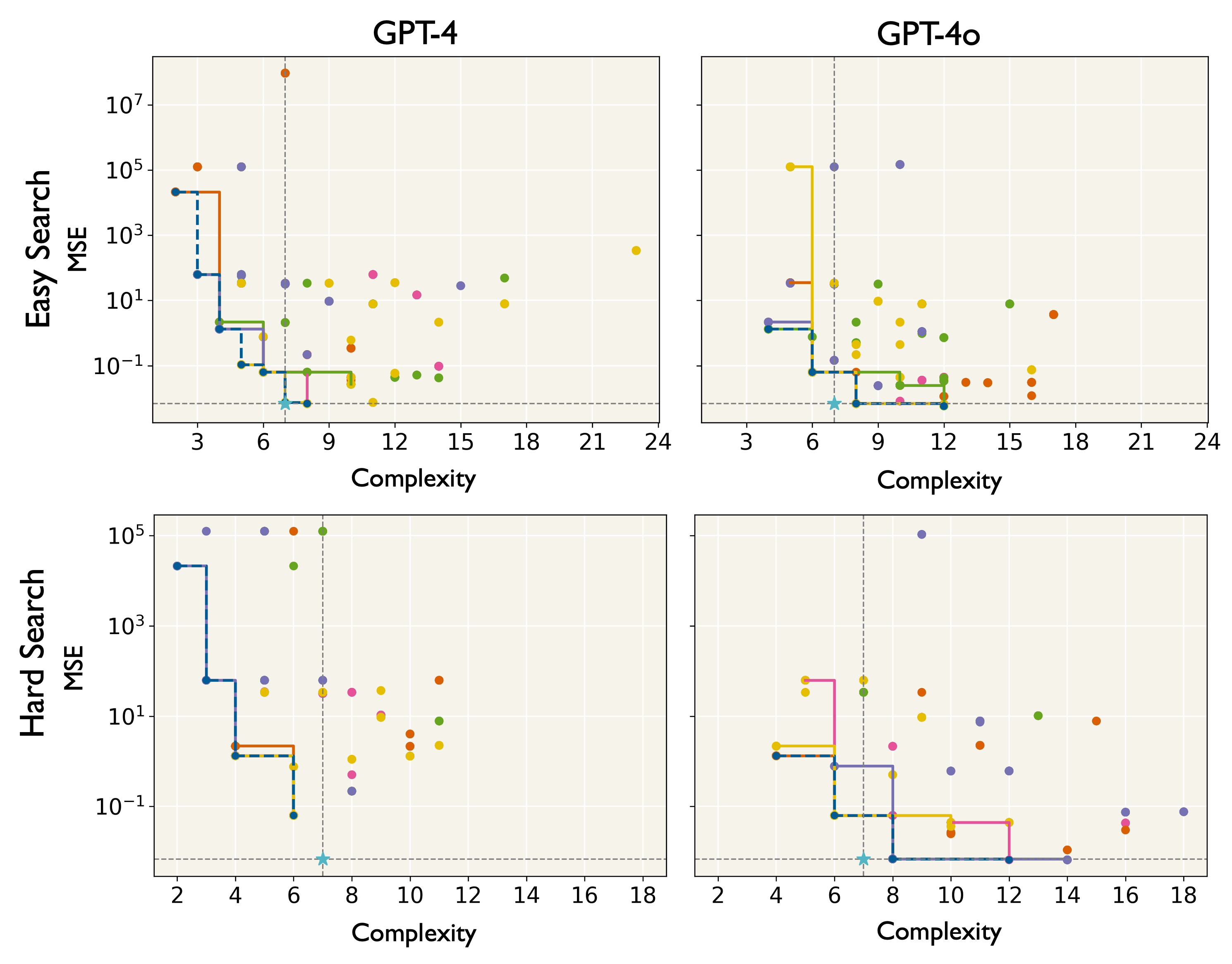}
    \caption
    {\textbf{Pareto fronts for Bode's Law.} }
\label{bode_paretoFronts}
\end{figure}

\begin{figure}[H]
    \centering
    \includegraphics[width=0.8\textwidth]{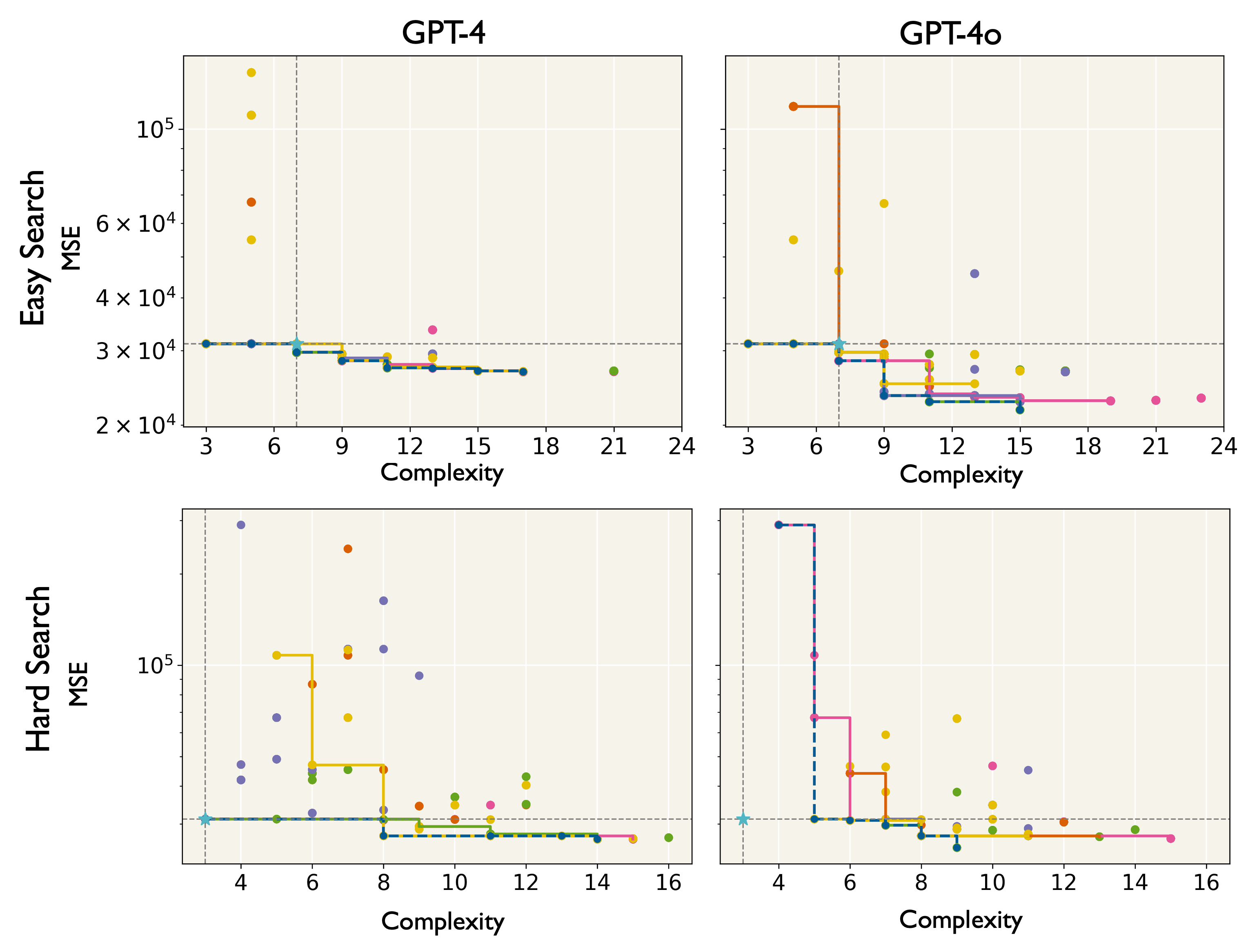}
    \caption
    {\textbf{Pareto fronts for Hubble's Law.} }
\label{hubble_paretoFronts}
\end{figure}

\begin{figure}[H]
    \centering
    \includegraphics[width=0.8\textwidth]{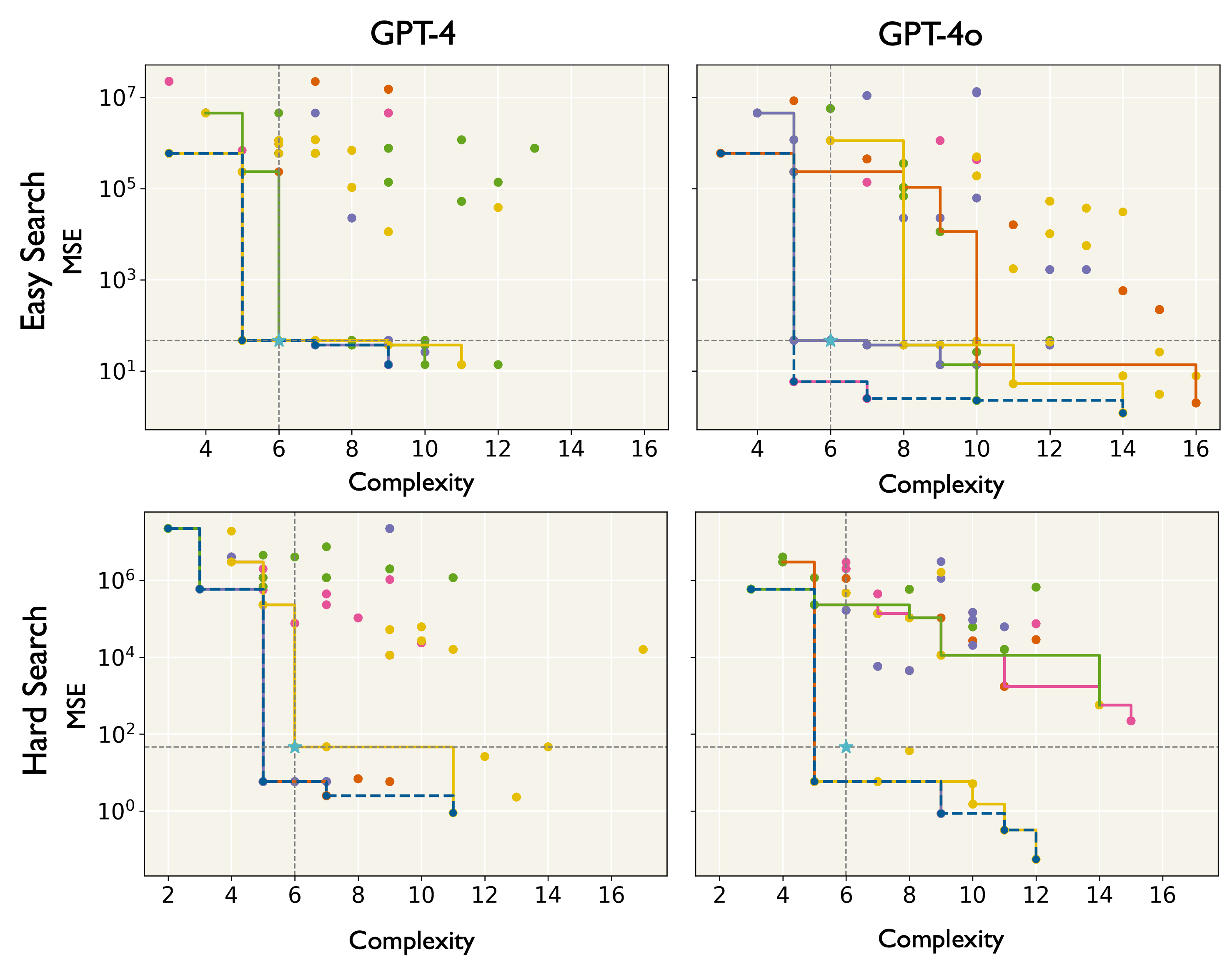}
    \caption
    {\textbf{Pareto fronts for Kepler’s Third Law.} }
\label{kepler_paretoFronts}
\end{figure}

\begin{table}[H]
\begin{center}
\begin{tabular}{lll}
\rowcolor[HTML]{EFEFEF} Expression & Mean Absolute Error (MAE) & Complexity \\
\rowcolor[HTML]{EFEFEF} &  &  \\
\rowcolor[HTML]{EFEFEF} &  &  \\
\specialrule{.1em}{.1em}{.1em} 
P1S1  & $0.02270419$ & $13$ \\
P1S2  & $0.00978477$ & $29$ \\
P2S1  & $0.00897093$ & $69$ \\
P2S2  & $0.00931620$ & $49$ \\
P3S1  & $0.01086397$ & $41$ \\
P3S2  & $0.00992712$ & $49$ \\
P1S1o & $0.02007803$ & $19$ \\
P1S2o & $0.02141686$ & $17$ \\
P2S1o & $0.00954461$ & $27$ \\
P2S2o & $0.01186963$ & $27$ \\
P3S1o & $0.00923655$ & $27$ \\
P3S2o & $0.01144178$ & $19$ \\
\specialrule{.1em}{.1em}{.1em}
\\
\end{tabular}
\end{center}
\caption{\textbf{Expressions of LLM models for Nikuradse dataset with complexity and MAE.} Here, P refers to prompt versions and S refers to dataset. The expressions generated by the GPT4o model are marked with an `o' at the end.} \label{NikExpTab}
\end{table}

\newpage
\bibliographystyle{unsrt}
\bibliography{references_revised}

\end{document}